\documentclass[nohyperref]{article}

\usepackage{microtype}
\usepackage{graphicx}
\usepackage{subfigure}
\usepackage{booktabs} % for professional tables

\usepackage{amsmath,amsfonts,bm}

\def\eqref#1{equation~\ref{#1}}
\def\1{\bm{1}}

\def\vx{{\bm{x}}}

\DeclareMathAlphabet{\mathsfit}{\encodingdefault}{\sfdefault}{m}{sl}
\SetMathAlphabet{\mathsfit}{bold}{\encodingdefault}{\sfdefault}{bx}{n}

\usepackage{hyperref}

\usepackage[accepted]{icml2022}

\usepackage{amsmath}
\usepackage{amssymb}
\usepackage{mathtools}
\usepackage{amsthm}

\usepackage[capitalize,noabbrev]{cleveref}

\theoremstyle{plain}

\theoremstyle{definition}

\theoremstyle{remark}

\usepackage[textsize=tiny]{todonotes}

\icmltitlerunning{Benchmarking Robustness of 3D Point Cloud Recognition Against Common Corruptions}

\begin{document}

\twocolumn[
\icmltitle{Benchmarking Robustness of 3D Point Cloud Recognition \\ Against Common Corruptions}
% On Benchmarking Robustness of 3D Point Cloud Recognition against Out-of-Distribution Shifts
% Benchmarking Robustness of  3D Point Cloud Recognition to Common Corruptions



\icmlsetsymbol{equal}{*}

\begin{icmlauthorlist}
\icmlauthor{Jiachen Sun}{yyy}
\icmlauthor{Qingzhao Zhang}{yyy}
\icmlauthor{Bhavya Kailkhura}{comp}
\icmlauthor{Zhiding Yu}{sch}
\icmlauthor{Chaowei Xiao}{sch,xxx}
\icmlauthor{Z. Morley Mao}{yyy}
\end{icmlauthorlist}

\icmlaffiliation{yyy}{University of Michigan, Ann Arbor}
\icmlaffiliation{comp}{Lawrence Livermore National Laboratory}
\icmlaffiliation{sch}{NVIDIA}
\icmlaffiliation{xxx}{ASU}

\icmlcorrespondingauthor{Jiachen Sun}{jiachens@umich.edu}

\icmlkeywords{Machine Learning, ICML}

\vskip 0.3in
]



\printAffiliationsAndNotice{}  % leave blank if no need to mention equal contribution

\begin{abstract}
Deep neural networks on 3D point cloud data have been widely used in the real world, especially in safety-critical applications. However, their robustness against corruptions is less studied. In this paper, we present ModelNet40-C, the first comprehensive benchmark on 3D point cloud \textit{corruption robustness}, consisting of 15 common and realistic corruptions. Our evaluation shows a significant gap between the performances on ModelNet40 and ModelNet40-C for state-of-the-art (SOTA) models. To reduce the gap, we propose a simple but effective method by combining PointCutMix-R and TENT after evaluating a wide range of augmentation and test-time adaptation strategies.
We identify a number of critical insights for future studies on corruption robustness in point cloud recognition. For instance, we unveil that Transformer-based architectures with proper training recipes achieve the strongest robustness. We hope our in-depth analysis will motivate the development of robust training strategies or architecture designs in the 3D point cloud domain. Our codebase and dataset are included in \url{https://github.com/jiachens/ModelNet40-C}.
\end{abstract}

\vspace{-0.75cm}
\section{Introduction}
\label{sec:intro}

Point clouds are one of the most acknowledged data format in 3D computer vision tasks, as they are inherently flexible representations and can be retrieved from a variety of sensors and computer-aided design (CAD) models. Because of these strengths, point clouds have been increasingly utilized in real-world applications, particularly in safety-critical areas like self-driving cars~\cite{yin2021center}, robotics~\cite{pomerleau2015review}, medical imaging~\cite{wang2019deep}, and virtual and augmented reality (VR/AR)~\cite{maloca2018high}. 
Processing of point clouds is thus crucial under these circumstances. For instance, autonomous vehicles rely on correct recognition of LiDAR point clouds to perceive the surroundings. Similar to 2D image classification, recent efforts demonstrate that deep learning models has dominated the point cloud recognition task.

As opposed to stellar progress on model architectures in 2D computer vision, deep 3D point cloud recognition is emerging where various architectures and operations are being proposed. Classic approaches discretize the point cloud into 3D cells, which causes cubic complexity. PointNet~\cite{qi2017pointnet} innovates to achieve end-to-end learning on point clouds. A few studies optimize the convolutional operation to be preferable for 3D point cloud learning~\cite{wang2019dynamic,liu2019relation}. Transformer~\cite{NIPS2017_3f5ee243} blocks are also applied as backbones in point cloud recognition architectures~\cite{guo2021pct}. The most extensively utilized benchmark for comparing methods for point cloud recognition is ModelNet40~\cite{wu20153d}. Although the accuracy on ModelNet40 over the past several years has been steadily improved, it merely shows a single perspective of model performance on the clean data. 
Given the importance of 3D point cloud in the safety-critical application, a comprehensive \emph{robustness} benchmark for point cloud recognition models is necessary.

In the literature, the vast majority of research on robustness in 3D point cloud recognition has concentrated on the critical difficulties of robustness against adversarial examples. Adversarial training has been adapted to defend against various threats to point cloud learning~\cite{sun2020adversarial1,sun2021adversarially}. 
% Recent studies also present certified defenses against adversarial examples~\cite{liu2021pointguard}. 
However, we find that the inevitable sensor inaccuracy and physical constraints will result in a number of \textit{common corruption} on point cloud data. For example, occlusion is a typical corruption for LiDAR and other scanning devices, rendering partially visible point clouds. Deformation is also ubiquitous in AR/VR games. Such corruptions pose a even bigger threat in most real-world application scenarios. Therefore, it is imperative to study the corruption robustness of 3D point cloud recognition. 

% Moreover, it is shown that data augmentation strategies like AugMix \cite{hendrycks2019augmix} and AutoAugment \cite{cubuk2018autoaugment} can improve the corruption robustness in 2D vision tasks. Apart from the architectural diversity in
% point cloud recognition models, a number of data augmentation and regularization techniques~\cite{zhang2021pointcutmix,chen2020pointmixup} have been proposed to improve the general performance of point cloud classification. It is, nevertheless, left unexplored how these methods will contribute to the corruption robustness of point cloud recognition. 
\begin{figure}[t]
\vspace{-0.1cm}
\includegraphics[width=0.975\linewidth]{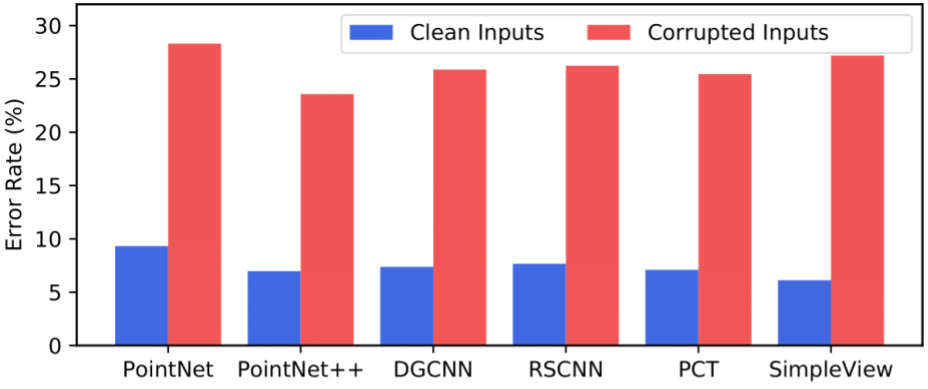}
\vspace{-0.4cm}
\caption{Despite the impressive results on clean inputs (\textit{i.e.}, ModelNet40), state-of-the-art point cloud recognition models cannot deliver good performance on corrupted inputs (\textit{i.e.}, ModelNet40-C). The error rate is $3\times$ larger on ModelNet40-C than ModelNet40.}\label{fig:teaser}
\vspace{-0.6cm}
\end{figure}

\textbf{Summary of Our Contributions}:

In this paper, we create, to our knowledge, the {\em first} systematic corruption robustness benchmark,  ModelNet40-C, for 3D point cloud recognition and present an in-depth analysis. To construct the dataset, we meticulously design and formulate 75 corruptions (15 types with 5 severity levels) that cover the majority of real-world point cloud distortion cases. We further provide a taxonomy of these corruptions into three categories (\textit{i.e}, density, noise and transformation)  and discuss their application scenarios. We anticipate that ModelNet40-C will serve as a first step towards 3D point cloud corruption-resistant models.

We conduct extensive evaluation on our ModelNet40-C. Specifically, we compare six representative models including \ PointNet~\cite{qi2017pointnet}, PointNet++~\cite{qi2017pointnet++}, DGCNN~\cite{wang2019dynamic}, RSCNN~\cite{liu2019relation}, PCT~\cite{guo2021pct}, and SimpleView~\cite{goyal2021revisiting}. We find that current models are vulnerable to our created corruptions. 
As shown in Fig.~\ref{fig:teaser}, there are nearly $3\times$ error rate gaps between model performances on ModelNet40 and ModelNet40-C. Our results reveal that \textbf{\textit{there is still considerable room for point cloud recognition models to improve on robustness against common corruptions}}.

To mitigate such gaps, we propose a simple but effective strategy by combing PointCutMix-R and TENT, after evaluating a wide range of data augmentation and test-time adaptation methods. Our method on average achieves the lowest error rate of 15.4\%. Specifically, we try augmentation (or regularization) strategies including  PointCutMix-R, PointCutMix-K, PointMixup, RSMix, and adversarial training based strategy. Additionally, we employ test-time adaptation methods (\textit{i.e.}, BN~\cite{schneider2020improving} and TENT~\cite{wang2020tent}) to show their potential in improving corruption robustness. We examine \textit{every} feasible combinations of architectures and methodologies, a total of 3,180 different configurations of experiments.

We summarize four conclusive insights below and eleven detailed findings in \S~\ref{sec:exp}. 

\textbullet~\textit{Insight 1}. Occlusion corruptions, rotation transformation, and background noises pose significant challenges for most point cloud recognition models (\S~\ref{sec:exp_corrup}). 

\textbullet~\textit{Insight 2}. 
Different architectures are vulnerable to different corruption types, which can be attributed to their design principles. (\S~\ref{sec:exp_arch}).
% Different architectures are more vulnerable to certain sorts of corruptions, and such results are consistent with their design principles. (\S~\ref{sec:exp_arch}).

\textbullet~\textit{Insight 3}. Different data augmentation strategies are especially advantageous for certain types of corruptions, which also correlate well with their design choices (\S~\ref{sec:exp_aug}).
 
\textbullet~\textit{Insight 4}. Test-time adaptations (BN and TENT) are beneficial for enhancing the corruption robustness, particularly for hard corruptions like occlusions and rotations  (\S~\ref{sec:exp_adapt}).

We hope our comprehensive benchmark and in-depth analysis will shed light on the robustness of point cloud recognition, and facilitate the development of robust architectures and training- and test-time robustness strategies on ModelNet40-C.

\vspace{-0.2cm}
\section{Related Work}
% In this section, we review a number of topics that relate to the corruption robustness of 3D point cloud perception.

\textbf{Adversarial \& Corruption Robustness of 2D Images}.
Deep neural networks are known to be vulnerable to adversarial examples and common corruptions~\cite{bulusu2020anomalous}. \citet{hendrycks2019benchmarking,hendrycks2021many} developed corruption robustness benchmarking datasets CIFAR-10/100-C, ImageNet-C, and ImageNet-R to facilitate robustness evaluations of CIFAR and ImageNet classification models. 
\citet{michaelis2019benchmarking} extended this benchmark to object detection models. 
\citet{mintun2021interaction} further proposed ImageNet-$\bar{\text{C}}$ dataset that is comprised of a set of corruptions that are perceptually dissimilar to ImageNet-C. 
Recently, \citet{sun2021certified} proposed a comprehensive benchmarking suite CIFAR-10/100-F that contains corruptions from different regions in the spectral domain. \cite{koh2021wilds} presented WILDS, a curated benchmark of $10$ datasets reflecting a diverse range of distribution shifts that naturally arise in real-world applications.~\citet{hendrycks2019augmix,cubuk2018autoaugment,calian2021defending} proposed augmentation methods to improve the corruption robustness in 2D vision tasks. On the adversarial robustness benchmarking front, \citet{carlini2019evaluating} discussed the methodological foundations, reviewed commonly accepted best practices, and suggested new methods for evaluating defenses to adversarial examples. \citet{croce2020robustbench} proposed a standardized leaderboard called RobustBench, which evaluates the adversarial robustness with AutoAttack~\cite{croce2020reliable}, a comprehensive ensemble of white- and black-box attacks. 
% Some other efforts in this direction are \cite{robustness} and \cite{dong2020benchmarking}.

\textbf{3D Point Cloud Deep Learning}. Deep learning models are increasingly being proposed to process point cloud data. Early works attempted to use 3D voxel grids for perception, which have cubic complexity~\cite{maturana2015voxnet,wang2015voting}. PointNet~\cite{qi2017pointnet} pioneered to leverage shared multi-layer perceptrons and a global pooling operation to achieve permutation-invariance and thus enable end-to-end training. \citet{qi2017pointnet++} further proposed PointNet++ to hierarchically stack PointNet for multi-scale local feature encoding. PointCNN and RSCNN refactor the traditional pyramid CNN to improve the local feature learning for point cloud recognition~\cite{li2018pointcnn,liu2019relation}. The graph data structure is also heavily used in point cloud learning~\cite{landrieu2018large,shen2018mining}. For example, DGCNN built a dynamic graph of point cloud data for representation learning~\cite{wang2019dynamic}. PointConv and KPConv improve the convolution operation for point cloud learning~\cite{wu2019pointconv,thomas2019kpconv}. Recent work demonstrated that ResNet~\cite{he2016deep} on multi-view 2D projections of point clouds could also achieve high accuracy~\cite{goyal2021revisiting}. PointTransformer and PCT advance Transformer~\cite{NIPS2017_3f5ee243} blocks into point cloud learning and achieve state-of-the-art performance~\cite{zhao2021point,guo2021pct}.

\textbf{Robustness Enhancements for 3D Point Cloud}. Several recent efforts tackle improving the robustness of 3D point cloud learning~\cite{sun2020towards}.~\citet{xiang2019generating} and~\citet{liu2019extending} first demonstrated that point cloud recognition is vulnerable to adversarial attacks. \citet{zhou2019dup} and \citet{dong2020self} proposed to leverage input randomization techniques to mitigate such vulnerabilities.~\citet{sun2020adversarial1} conducted adaptive attacks on existing defenses and analyzed the application of adversarial training on point cloud recognition.~\citet{zhao2020isometry} discovered that adversarial rotation greatly degrades the perception performance.~\citet{sun2021adversarially} further showed that pre-training on self-supervised tasks enhances the adversarial robustness of point cloud recognition. Recent studies presented a framework that uses the Shapley value~\cite{roth1988shapley} to assess the quality of representations learned by different point cloud recognition models~\cite{shen2021interpreting,Shen_2021_CVPR}. Recent efforts also proposed certified adversarial defenses\cite{liu2021pointguard}. However, little attention has been paid to the common corruption robustness of point cloud recognition. In this work, we aim to present a systematic benchmark and rigorously analyze the corruption robustness of representative deep point cloud recognition models.

\begin{figure*}[t]
\vspace{-0.1cm}
\includegraphics[width=1\linewidth]{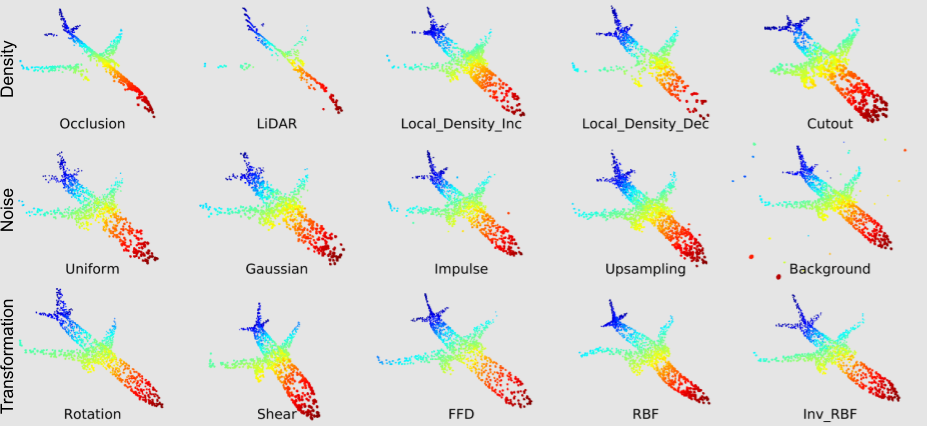}
\vspace{-0.5cm}
\caption{Visualizations of Our Constructed ModelNet40-C. Our ModelNet40-C dataset consists of 15 corruption types that represent different out-of-distribution shifts in real-world applications of point clouds. Similar to ImageNet-C~\cite{hendrycks2019benchmarking}, each corruption type has five severity levels. We carefully examine the generated point clouds and ensure they preserve their original semantics. More visualization samples are shown in Appendix~\ref{ap:modelnet}}\label{fig:modelnetc}
\vspace{-0.5cm}
\end{figure*}
\vspace{-0.2cm}
\section{3D Point Cloud Corruption Robustness}
\label{sec:cor}

In this section, we introduce the design principles of our 3D corruption benchmark. As mentioned in \S~\ref{sec:intro}, 3D point clouds are being utilized in various safety- and security-critical real-world applications~\cite{Geiger2012CVPR,adallah20193d,yu20213d,cheng2020morphing}. Extensive studies have been carried out to improve both architectures and training strategies for point cloud recognition on in-distribution data~\cite{qi2017pointnet,wang2019dynamic,chen2020pointmixup,lee2021regularization}. However, there has not been any systematic study on the model robustness against common corruption. 
To bridge this gap, we design 15 common corruptions for benchmarking \textit{corruption robustness} of point cloud recognition models. It is worth noting that such designs are \textit{non-trivial} since the manipulation space of 3D point clouds is completely different from 2D images where the corruptions come from the RGB modification~\cite{hendrycks2019benchmarking}. In particular, we have three principles to design our benchmarks: 
i) Since we directly manipulate the position of points, we need to take extra care to preserve the \textit{original semantics} of point clouds (Fig.~\ref{fig:modelnetc}). ii) we should ensure the constructed corruptions are \textit{realistic} in various applications. iii) We should take \textit{diversity} as an important factor to emulate a wide range of natural corruptions for 3D point clouds.

Our 15 corruption types can be naturally grouped into three categories (\textit{i.e.}, density, noise, and transformation) , and we will introduce them in the following subsections.   

\vspace{-0.2cm}
\subsection{Density Corruption Patterns}

3D point cloud can be collected from various sensors like VR scanning devices and LiDAR for autonomous driving, or generated from computer-aided design (CAD) models. Therefore, the testing point clouds may have different density patterns from the training samples. For example, VR scanning (in indoor scenes) and LiDAR sensors may suffer from occlusion, so that only a portion of the point cloud is visible~\cite{Geiger2012CVPR,dai2017scannet}. Besides, the direct reflection of lasers on metal materials will cause local missing points in LiDAR point clouds~\cite{liu2018tof}. The local density of 3D scanned point clouds rely on how frequently the device passes that area~\cite{nguyen20133d}. We hence formulate five corruption types to cover the density corruption patterns: \{\texttt{Occlusion}, \texttt{LiDAR}, \texttt{Local\_Density\_Inc}, \texttt{Local\_Density\_Dec}, \texttt{Cutout}\}. Specifically, \texttt{Occlusion} and \texttt{LiDAR} both simulate occlusion patterns using ray tracing on the original meshes~\cite{Zhou2018}, and \texttt{LiDAR} additionally incorporates the vertically line-styled pattern of LiDAR point clouds~\cite{liu2018tof}. \texttt{Local\_Density\_Inc} and \texttt{Local\_Density\_Dec} will randomly select several local clusters of points using $k$-nearest neighbors ($k$NN) to increase and decrease their density, respectively. Similarly, \texttt{Cutout} discards several randomly chosen local clusters of points using $k$NN~\cite{devries2017improved}.

\vspace{-0.2cm}
\subsection{Noise Corruption Patterns}

Noise evidently exists in all real-world point cloud applications. For example, the inevitable digital noise of scanning sensors (\textit{e.g.}, medical imaging)~\cite{wolff2016point} and the random reflections and inaccuracy of LiDAR lasers~\cite{Geiger2012CVPR} will contribute to a substantial variation of points. Compression and decompression will potentially result in noisy point clouds as well~\cite{cao20193d}. Besides, real-time rendering in VR games is another source of noise~\cite{bonatto2016explorations}. Although noise is a common corruption pattern for both 2D and 3D data, the manipulation space is larger for point clouds since their numbers of points are adjustable. We thus formulate five noise perturbations: \{\texttt{Uniform}, \texttt{Gaussian}, \texttt{Impulse}, \texttt{Upsampling}, \texttt{Background}\}. As their names indicate, \texttt{Uniform} and \texttt{Gaussian} apply different distributional noise to each point in a point cloud. \texttt{Impulse} applies deterministic perturbations to a subset of points. \texttt{Upsampling} assigns new perturbation points around the existing points. \texttt{Background} randomly adds new points in the bounding box space of the pristine point cloud.

\vspace{-0.2cm}
\subsection{Transformation Corruptions Patterns}

We use both linear and non-linear 3D transformations to formulate the corruptions. For the linear ones, we leverage 3D \texttt{Rotation} and \texttt{Shear} as our corruption types and exclude translation and scale transformations since they can be easily restored by normalization (\textit{i.e.}, the inverse transformation matrix). \texttt{Rotation} of point clouds is common in the real world and the robustness against adversarial rotations has been investigated by a few studies~\cite{zhao2020isometry,shen2021interpreting}. We here do not use aggressive rotations that might affect human perception as well, but instead enable a milder rotation ($\le 15^{\circ}$) along $xyz$ axes. We consider \texttt{Shear} on the $xy$ plane to represent the motion distortion in 3D point clouds~\cite{yang2021lidar}. We utilize free-form deformation (FFD)~\cite{sederberg1986free} and radial basis function (RBF)-based deformation~\cite{forti2014efficient} for non-linear transformations. Such deformations are also common in VR/AR games and point clouds from generative models (GAN)~\cite{li2018point,zhou20213d}. Specifically, we use multi quadratic ($\varphi(\boldsymbol{x}) = \sqrt{\boldsymbol{x}^2 + r^2}$) and inverse multi quadratic splines ($\varphi(\boldsymbol{x}) = (\boldsymbol{x}^2 + r^2 )^{-\frac{1}{2}}$) as the representative RBFs to cover a wide range of deformation types. As a result, we in total formulate \{\texttt{Rotation}, \texttt{Shear}, \texttt{FFD}, \texttt{RBF}, \texttt{Inv\_RBF}\} as our transformation-based corruptions. 
\vspace{-0.3cm}
\section{ModelNet40-C Robustness Benchmark}

ModelNet40 is the most popular dataset for benchmarking point cloud recognition performance, containing 12,308 point clouds from 40 classes~\cite{wu20153d}. Point clouds from ModelNet40 are extracted from CAD models, rendering a perfectly clean dataset. \citet{uy2019revisiting} recently proposed ScanObjectNN consisting of point clouds scanned from real-world objects to show the performance gap between models trained on synthetic and real-world data. However, quantifying how models trained on the clean dataset perform under common corruptions encountered during the test time remains challenging. To this end, we use the ModelNet40 as our base dataset to construct ModelNet40-C. It is worth noting that our devised corruptions are general to be applied to other dataset, like ShapeNet~\cite{chang2015shapenet}.

\textbf{Setup}. We create ModelNet40-C with five severity levels for each corruption type, the same as ImageNet-C. Fig.~\ref{fig:modelnetc} illustrates samples from ModelNet40-C with severity level four, and they clearly still preserve the semantics of the ``airplane'' class.  Since it is hard to qualify and quantify the corruption severity for \texttt{LiDAR} and \texttt{Occlusion}, we instead leverage five different view angles to create their corrupted point clouds. The detailed construction of ModelNet40-C is introduced in Appendix~\ref{ap:modelnet}. These designed corruptions are applied to the \textit{validation} set of ModelNet40, resulting in ModelNet40-C a 75$\times$ larger dataset to test the corruption robustness of pre-existing models. \textit{Note that ModelNet40-C should be only used in the test time rather than in the training phase.}

\textbf{Metrics}. We use the error rate (ER) and class-wise mean error rate (mER) as the main metrics for ModelNet40-C benchmarking. We denote $\text{ER}_\text{clean}^f$ as the error rate for a classifier $f$ on the clean dataset (\textit{i.e.}, ModelNet40) and $\text{ER}_{s,c}^f$ as the error rate for $f$ on corruption $c$ with severity $s$. Similarly, $\text{ER}_{c}^{f}=\sum_{s=1}^{5} \text{ER}_{s,c}^f$ and $\text{ER}_\text{cor}^{f}=\sum_{c=1}^{15} \text{ER}_\text{c}^f$. We will release our leaderboard publicly to facilitate future studies on robustness of point cloud learning. The goal of ModelNet40-C is to evaluate the general robustness of point cloud learning models in various real-world scenarios. ModelNet40-C is not designed for training-time optimizations but augmentations with other/similar corruptions are allowed and should be explicitly stated. The corruption robustness of trained models can be assessed by their performance on ModelNet40-C using above metrics.

\begin{table}[t]
\scriptsize
\renewcommand\arraystretch{1.}
\setlength\tabcolsep{2pt}
  \vspace{-0.3cm}
  \caption{\small Error Rates of Different Models with Training Strategies on the ModelNet40 ($\text{ER}_\text{clean}$). \textbf{Bold} and \underline{underline} denote the best and runner-up results throughout this paper, respectively.
%   \AM{Missing number might draw unnecessary criticism. Is there a better way to combine the results?}
  %\BK{Better caption with message/results -- eg see winning hand captions}\AM{Move the detailed result to Appendix and only keep first 5 columns.}
  }
  \label{tb:clean}
  \centering
\resizebox{1.\linewidth}{!}{
  \begin{tabular}{l||c|ccccc}
    \specialrule{1pt}{1.1pt}{1pt}
    %  &CIFAR-10 &\multicolumn{4}{c|}{CIFAR-10-C} &CIFAR-10-$\bar{\text{C}}$ &CIFAR-100 &\multicolumn{4}{c|}{CIFAR-100-C} &CIFAR-100-$\bar{\text{C}}$ &ImageNet & \multicolumn{4}{c|}{ImageNet-C} &ImageNet-$\bar{\text{C}}$ \\
    % \hline
    %Pretext Task     & Parameters  &Clean Accuracy &Adv. Accuracy  &Clean Accuracy &Adv. Accuracy  &Clean Accuracy &Adv. Accuracy\\
    Model (\%) $\downarrow$   &Standard & PointCutMix-R & PointCutMix-K & PointMixup & RSMix &PGD\\
\noalign{\global\arrayrulewidth1pt}\hline\noalign{\global\arrayrulewidth0.4pt}     
PointNet   &  9.3 &  9.4 &  9.0 &  8.9 & 9.8 &  11.8 \\
PointNet++ &  \underline{7.0} &  \textbf{7.1} &  \textbf{6.7} &  \textbf{7.1} &  \textbf{6.6} &  - \\
DGCNN      &  7.4 &  7.4 &  \underline{6.8} &  7.8 &  \underline{7.1} &  \textbf{8.1} \\
RSCNN      &  7.7 &  7.6 &  7.1 &  \underline{7.2} &  7.6 &  - \\
PCT        &  7.1 &  \underline{7.2} &  6.9 &  7.4 &  6.9 &  \underline{8.9} \\
SimpleView &  \textbf{6.1} &  7.9 &  7.4 &  \underline{7.2} &  7.9 &  - \\
\noalign{\global\arrayrulewidth1pt}\hline\noalign{\global\arrayrulewidth0.4pt}
  \end{tabular}
  }
  \vspace{-0.5cm}
\end{table}
\vspace{-0.3cm}
\section{Experiments}
\label{sec:exp}

\begin{table*}[t]
\scriptsize
\renewcommand\arraystretch{1.05}
\setlength\tabcolsep{2pt}
\vspace{-0.25cm}
  \caption{\small Error Rates of Different Model Architectures on ModelNet40-C with Standard Training. 
%   \AM{Missing number might draw unnecessary criticism. Is there a better way to combine the results?}
  %\BK{Better caption with message/results -- eg see winning hand captions}\AM{Move the detailed result to Appendix and only keep first 5 columns.}
  }
  \label{tb:total1}
  \centering
 \resizebox{1.03\textwidth}{!}{
  \begin{tabular}{l||c|ccccc|ccccc|ccccc}
  \specialrule{1pt}{1.1pt}{1pt}
    & & \multicolumn{5}{c|}{Density Corruptions} & \multicolumn{5}{c|}{Noise Corruptions} & \multicolumn{5}{c}{Transformation Corruptions}\\
    \cline{3-17}
    %  &CIFAR-10 &\multicolumn{4}{c|}{CIFAR-10-C} &CIFAR-10-$\bar{\text{C}}$ &CIFAR-100 &\multicolumn{4}{c|}{CIFAR-100-C} &CIFAR-100-$\bar{\text{C}}$ &ImageNet & \multicolumn{4}{c|}{ImageNet-C} &ImageNet-$\bar{\text{C}}$ \\
    % \hline
    %Pretext Task     & Parameters  &Clean Accuracy &Adv. Accuracy  &Clean Accuracy &Adv. Accuracy  &Clean Accuracy &Adv. Accuracy\\
    Model (\%) $\downarrow$ &$\text{ER}_\text{cor}$ & Occlusion & LiDAR & Density Inc. & Density Dec.  &Cutout & Uniform &Gaussian &Impulse &Upsampling  &Background  &Rotation &Shear &FFD &RBF  &Inv. RBF\\
\noalign{\global\arrayrulewidth1pt}\hline\noalign{\global\arrayrulewidth0.4pt}     
PointNet   & 28.3 & \underline{52.3} & \textbf{54.9} & \textbf{10.5} & \underline{11.6} & \underline{12.0} & \underline{12.4} & 14.4 & 29.1 & \textbf{14.0} & 93.6 & 36.8 & 25.4 & 21.3 & 18.6 & 17.8 \\
PointNet++ & \textbf{23.6} & 54.7 & \underline{66.5} & 16.0 & \textbf{10.0} & \textbf{10.7} & 20.4 & 16.4 & 35.1 & \underline{17.2} & \underline{18.6} & 27.6 & 13.4 & 15.2 & 16.4 & 15.4 \\
DGCNN      & 25.9 & 59.2 & 81.0 & 14.1 & 17.3 & 15.4 & 14.6 & 16.6 & \underline{24.9} & 19.1 & 53.1 & \underline{19.1} & \underline{12.1} & \underline{13.1} & \underline{14.5} & \underline{14.0} \\
RSCNN      & 26.2 & \textbf{51.8} & 68.4 & 16.8 & 13.2 & 13.8 & 24.6 & 18.3 & 46.2 & 20.1 & \textbf{18.3} & 29.2 & 17.0 & 18.1 & 19.2 & 18.6 \\
PCT        & \underline{25.5} & 56.6 & 76.7 & \underline{11.8} & 14.3 & 14.5 & \textbf{12.1} & \textbf{13.9} & 39.1 & 17.4 & 57.9 & \textbf{18.1} & \textbf{11.5} & \textbf{12.4} & \textbf{13.0} & \textbf{12.6} \\
SimpleView & 27.2 & 55.5 & 82.2 & 13.7 & 17.2 & 20.1 & 14.5 & \underline{14.2} & \textbf{24.6} & 17.7 & 46.8 & 30.7 & 18.5 & 17.0 & 17.9 & 17.2 \\
\hline
Average    & 26.1 & 55.0 & 71.6 & 13.8 & 13.9 & 14.4 & 16.4 & 15.6 & 33.2 & 17.6 & 48.0 & 26.9 & 16.3 & 16.2 & 16.6 & 15.9 \\
\noalign{\global\arrayrulewidth1pt}\hline\noalign{\global\arrayrulewidth0.4pt}
  \end{tabular}
    }
  \vspace{-0.3cm}
\end{table*}

In this section, we elaborate our comprehensive evaluation and rigorous analysis in detail. We benchmark corruption robustness of point cloud recognition from the perspectives of corruption types and model architectures. Moreover, we examine the effectiveness of data augmentations and test-time adaptation methods as mitigation solutions against common corruptions.

\textbf{Setup}. We leverage six representative model architectures: PointNet~\cite{qi2017pointnet}, PointNet++~\cite{qi2017pointnet}, DGCNN~\cite{wang2019dynamic}, RSCNN~\cite{liu2019relation}, PCT~\cite{guo2021pct}, and SimplView~\cite{goyal2021revisiting}. These six models stand for distinct architecture designs, and have achieved good accuracy on the clean dataset. They are also well-recognized by the 3D vision community, and have been extensively applied to complex tasks like semantic segmentation~\cite{nguyen20133d} and object detection~\cite{shi2019pointrcnn,shi2020pv}. As suggested by~\citet{goyal2021revisiting}, we adopt the same training strategy for all models. We utilize smoothed cross-entropy~\cite{wang2019dynamic} as the loss function as it has been demonstrated to improve the recognition performance. We take 1024 points as input size in the training phase and use the Adam optimizer~\cite{kingma2014adam} with the \texttt{ReduceLROnPlateau} scheduler implemented in PyTorch~\cite{paszke2019pytorch}. We train 300 epochs and pick the best performant model for our further evaluation and follow~\citet{wang2019dynamic} to use random translation and scaling as our default data augmentation. All training and testing experiments are done on a GeForce RTX 2080 GPU. We report the class-wise mean ER in Appendix~\ref{ap:experiment} due to space constraints. 

\begin{figure}[t]
\centering
\vspace{-0.15cm}
\includegraphics[height=0.7\linewidth, width=1.\linewidth]{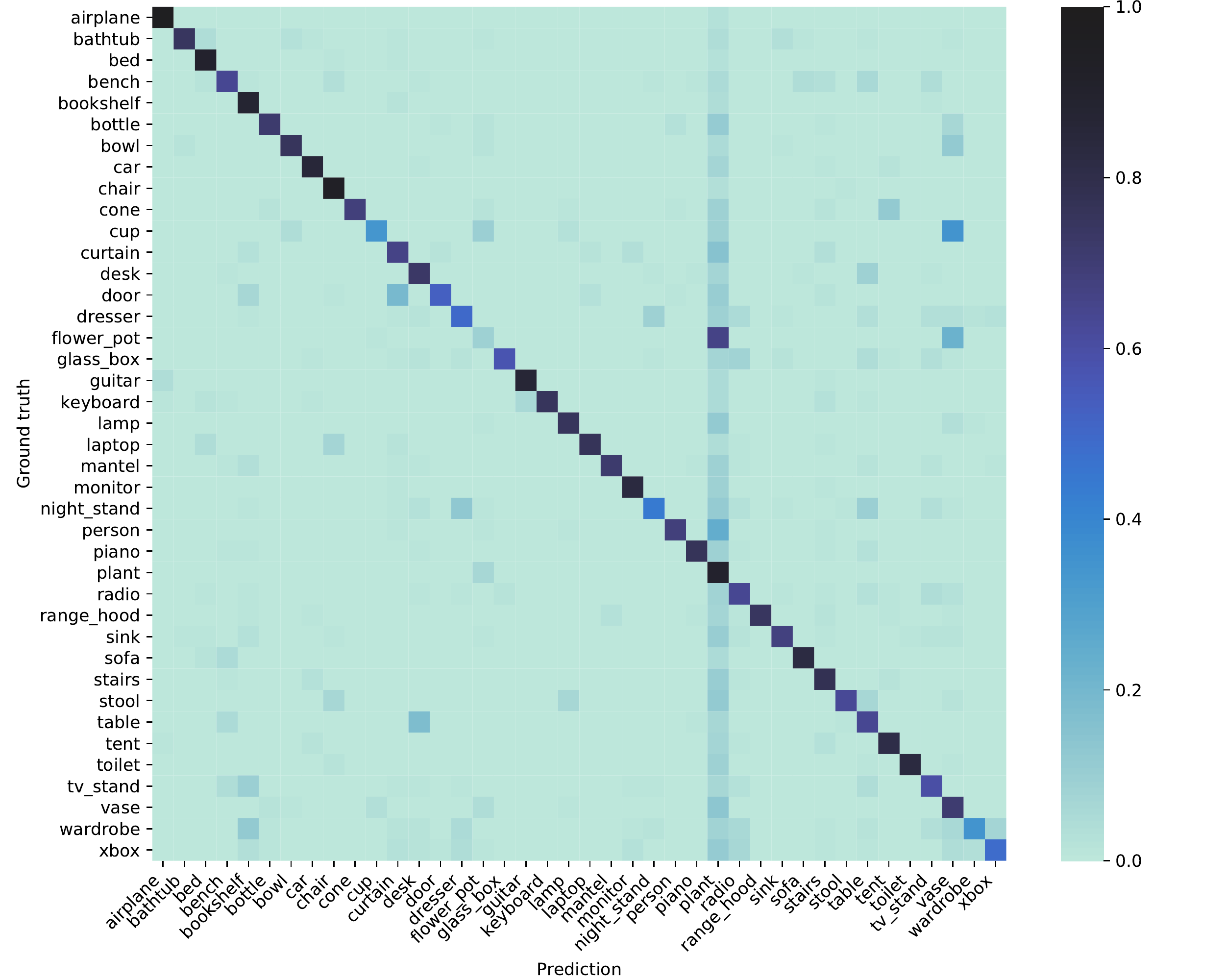}
\vspace{-0.65cm}
\caption{Confusion Matrix on ModelNet40-C Averaged over Six Models with Standard Training. Confusion matrices for each model are shown in Figure~\ref{fig:apconfusion} in Appendix~\ref{ap:experiment}.}\label{fig:confusion}
\vspace{-0.7cm}
\end{figure}
 
Besides, we try data augmentation and test-time adaption strategies and evaluate their effectiveness against our created corruptions. In particular, we leverage PointCutMix-R, PointCutMix-K \cite{zhang2021pointcutmix}, PointMixup~\cite{chen2020pointmixup}, RSMix~\cite{lee2021regularization}, and PGD-based adversarial training~\cite{sun2021adversarially} as additional data augmentation strategies. We adopt the original hyper-parameter settings from their official implementations in our study. Detailed introduction can be found in Appendix~\ref{ap:experiment}. We only enable adversarial training for PointNet, DGCNN, and PCT since PointNet++ and RSCNN leverage ball queries 
to find neighboring points and  SimpleView projects the pristine point cloud into multi-view 2D images. Both methods will hinder the gradients from backward propagating to the original point cloud, making adversarial training inapplicable. We leave them as future work. 

\textbf{Clean Performance}. Table~\ref{tb:clean} shows the $\text{ER}_\text{clean}$ of different model architectures with the adopted training strategies. All models achieve 90+\% accuracy with standard training. As~\citet{goyal2021revisiting} indicate, auxiliary factors will obscure the effect of architectures, and the performance gaps among different models are not significant on the ModelNet40 validation set. Data augmentation strategies like PointMixup claim to enhance the general model performance. However, with these factors controlled, we also do not find tangible improvements over these augmentation recipes. Besides, adversarially trained models are expected to perform slightly worse on the clean dataset compared to others~\cite{sun2021adversarially}. It also initiates that model performances on ModelNet40 tend to saturate. Thus, it is a necessary to evaluate the model effectiveness from other perspectives (\textit{e.g.}, robustness).

\subsection{Comparison Among Corruption Types}
\label{sec:exp_corrup}

\textbf{Validity}. We first benchmark robustness of six standard trained models to verify the validity of our ModelNet40-C. The detailed results are presented in Fig.~\ref{fig:erforda} and~\ref{fig:erforda2} in Appendix~\ref{ap:experiment}. We find that the $\text{ER}_{s,c}$ gradually increases as the severity level increasing for each corruption, which justifies our hyper-parameter setting. As shown in Fig.~\ref{fig:teaser}, there is a $\sim 3\times$ performance degradation between the overall $\text{ER}_\text{cor}$ and $\text{ER}_\text{clean}$ for all benchmarked models. Fig.~\ref{fig:confusion} shows the average confusion matrix over six models and 15 corruptions, where the values in the diagonal are still high, further validating the semantic maintenance of ModelNet40-C. 
ModelNet40-C fullfills its goal as a robustness benchmark for future studies.

Table~\ref{tb:total1} presents the detailed $\text{ER}_{c}$ of the six models evaluated on ModelNet40-C with standard training. As mentioned before, there is a significant increase in $\text{ER}_{c}$ on each corruption compared to $\text{ER}_\text{clean}$. The gap ranges from 17.6\% to 89.7\% among different corruptions. From the perspective of the corruption, we obtain several interesting observations.
% \textit{Occlusion}, \textit{LiDAR}, and \textit{Rotation} will degrade . \textit{Background} and \textit{Impulse}

\textbullet~\textit{Insight 1.1.} \texttt{Occlusion} and \texttt{LiDAR} corruptions pose a major threat for 3D point cloud recognition models.

From Table~\ref{tb:total1}, $\text{ER}_\text{Occlusion}$ and $\text{ER}_\text{LiDAR}$ reach 55.0\% and 71.6\% on average, respectively. Occlusion happens in most real-world application of 3D point clouds. Moreover, we find the models have poor performance regardless of the occlusion directions, suggesting a general vulnerability of point cloud recognition.

\textbullet~\textit{Insight 1.2.} \texttt{Rotation} is still a challenging corruption for 3D point cloud recognition models even with small angles.

Rotation is a well-known threat for point cloud recognition by several recent studies~\cite{zhao2020isometry,shen2021interpreting}. Existing studies allow a rotation angle (\textit{e.g.}, $\ge 45^\circ$). However, such rotated point clouds confuse human perception without RGB information. In our study, we find that a small rotation ($\le 15^\circ$) still causes a high ER on point cloud recognition models ranging from 18.1\% to 36.8\%.

\textbullet~\textit{Insight 1.3.} \texttt{Impulse} and \texttt{Background} corruptions are surprisingly troublesome to 3D point cloud recognition. 

We find $\text{ER}_\text{Impulse}$ (33.2\%) and $\text{ER}_\text{Background}$ (48.0\%) are abnormally high for most architectures. Although they are even less perceptible than Gaussian and uniform noise since only a small portion of points are affected. However, the magnitudes of \texttt{Impulse} and \texttt{Background} noises are high, suggesting that a small portion of outliers will greatly affect point cloud recognition performance.

% Moreover, we find that the standard trained models are more resilient to \texttt{Uniform} and \texttt{Gaussian} than 2D image classifiers.

\subsection{Comparison Among Model Architectures}
\label{sec:exp_arch}

As presented in Table~\ref{tb:total1}, there is no overarching model that dominates our ModelNet40-C dataset, unlike robustness benchmarking in 2D vision~\cite{hendrycks2019benchmarking}. Point cloud recognition models have various designs and no consensus has been reached as deep learning in the 3D space is a relatively nascent field. The model performances on ModelNet40-C are found to be in good alignment with their design attributes.

\textbullet~\textit{Insight 2.1.} PointNet achieves strong performance on density corruptions, but fails on transformation corruptions. 

\begin{table*}[t]
\scriptsize
\renewcommand\arraystretch{1.05}
\setlength\tabcolsep{2pt}
\vspace{-0.25cm}
  \caption{\small Error Rates of Different Model Architectures on ModelNet40-C with Different Data Augmentation Strategies.
%   \AM{Missing number might draw unnecessary criticism. Is there a better way to combine the results?}
  %\BK{Better caption with message/results -- eg see winning hand captions}\AM{Move the detailed result to Appendix and only keep first 5 columns.}
  }
  \label{tb:aug}
  \centering
\resizebox{1.025\textwidth}{!}{
  \begin{tabular}{l||c|c|ccc|c|ccc|c|ccc|c|ccc|c|ccc}
    \specialrule{1pt}{1.1pt}{1pt}
    & Standard &\multicolumn{4}{c|}{PointCutMix-R} & \multicolumn{4}{c|}{PointCutMix-K} & \multicolumn{4}{c|}{PointMixup} & \multicolumn{4}{c|}{RSMix} & \multicolumn{4}{c}{PGD}\\
    \cline{2-22}
    %  &CIFAR-10 &\multicolumn{4}{c|}{CIFAR-10-C} &CIFAR-10-$\bar{\text{C}}$ &CIFAR-100 &\multicolumn{4}{c|}{CIFAR-100-C} &CIFAR-100-$\bar{\text{C}}$ &ImageNet & \multicolumn{4}{c|}{ImageNet-C} &ImageNet-$\bar{\text{C}}$ \\
    % \hline
    %Pretext Task     & Parameters  &Clean Accuracy &Adv. Accuracy  &Clean Accuracy &Adv. Accuracy  &Clean Accuracy &Adv. Accuracy\\
    Model (\%) $\downarrow$ & $\text{ER}_\text{cor}$ & $\text{ER}_\text{cor}$ & Density & Noise & Trans. & $\text{ER}_\text{cor}$ & Density & Noise & Trans. & $\text{ER}_\text{cor}$ & Density & Noise & Trans. & $\text{ER}_\text{cor}$ & Density & Noise & Trans. & $\text{ER}_\text{cor}$ & Density & Noise & Trans.\\
\noalign{\global\arrayrulewidth1pt}\hline\noalign{\global\arrayrulewidth0.4pt}     
PointNet   &28.3 & 21.8 & 30.5 & 18.0 & 16.9 & 21.3 & 26.8 & 21.8 & 15.4 & 25.4 & \textbf{28.3} & 28.9 & 19.0 & 22.5 & \textbf{24.8} & 27.3 & 15.5 & 25.9 & \textbf{28.8} & 28.4 & 20.5 \\
PointNet++  &\textbf{23.6} & 19.1 & 28.1 & 12.2 & 17.0 & 20.2 & \underline{26.3} & 16.9 & 17.3 & \textbf{19.3} & 30.8 & \textbf{14.3} & 12.9 & 23.3 & 27.0 & 19.3 & 23.7 &  - &  - &  - &  - \\
DGCNN   &25.9    & \underline{17.3} & 28.9 & 11.4 & \underline{11.5} & \underline{17.3} & 29.1 & \textbf{11.9} & \textbf{10.9} & 20.4 & 32.1 & 16.8 & \underline{12.3} & \underline{18.1} & 28.8 & \underline{13.0} & \textbf{12.6} & \underline{20.7} & 36.8 & \textbf{13.8} & \underline{11.5} \\
RSCNN   &26.2    & 17.9 & \textbf{25.0} & 13.0 & 15.8 & 21.6 & 28.3 & 19.0 & 17.6 & 19.8 & \underline{29.7} & \underline{15.5} & 14.1 & 21.2 & 26.8 & 17.4 & 19.3 &  - &  - &  - &  - \\
PCT     &\underline{25.5}    & \textbf{16.3} & \underline{27.1} & \textbf{10.5} & \textbf{11.2} & \textbf{16.5} & \textbf{25.8} & \underline{12.6} & \underline{11.1} & \underline{19.5} & 30.3 & 16.7 & \textbf{11.5} & \textbf{17.3} & \underline{25.0} & \textbf{12.0} & \underline{15.0} & \textbf{18.4} & \underline{29.3} & \underline{14.7} & \textbf{11.1} \\
SimpleView  &27.2 & 19.7 & 31.2 & \underline{11.3} & 16.5 & 20.6 & 29.1 & 15.6 & 17.0 & 21.5 & 32.7 & 17.1 & 14.8 & 20.4 & 28.4 & 14.6 & 18.3 &  - &  - &  - &  - \\
\hline
Average &26.1   & 18.7 & 28.5 & 12.7 & 14.8 & 19.6 & 27.6 & 16.3 & 14.9 & 21.0 & 30.6 & 18.2 & 14.1 & 20.5 & 26.8 & 17.3 & 17.4 &  - &  - &  - &  - \\
\noalign{\global\arrayrulewidth1pt}\hline\noalign{\global\arrayrulewidth0.4pt}
  \end{tabular}
    }
  \vspace{-0.3cm}
\end{table*}

PointNet does not encode local feature, and several publications have studies it from the perspectives of adversarial robustness~\cite{sun2021adversarially} and representation quality~\cite{shen2021interpreting}. Such a design has been regarded as a main drawback of PointNet. However, we find it robust against the variations in density. Table~\ref{tb:total1} presents that PointNet achieves an $\text{ER}$ of 28.3\% on density corruptions, and overall outperforms the runner-up by 11.7\%. Such results can be attributed to the locality of density corruptions. Compared to other models that embeds complex local features, PointNet is less sensitive to local changes of the input point cloud. On the other hand, PointNet indeed fails on other corruptions, rendering itself the worst performant model on average. Our analysis complements the existing understanding of PointNet, we believe the usage of PointNet should be determined by the application scenarios.

\textbullet~\textit{Insight 2.2.} Ball query-based clustering operation is robust against \texttt{Background} noise.

As mentioned in \S~\ref{sec:exp_corrup}, \texttt{Background} is a challenging corruption to point cloud recognition. However, we find that PointNet++ and RSCNN are specially robust against it, which have outperform other models by 70.7\%. We discover that the ball query of neighboring points is the key to such robustness. Compared to $k$NN that has deterministic $k$ points to cluster, ball query fixes the radius to reject faraway points in the bounding box space. This design helps models tackle the root cause of the \texttt{Background} corruption. 

\textbullet~\textit{Insight 2.3.} Transformer-based architectures are robust against transformation corruptions. 

Transformer~\cite{NIPS2017_3f5ee243} has recently reformed the 2D vision~\cite{dosovitskiy2021an}. PCT leverages multiple Transformer blocks as its backbone, which leverage self-attention modules to embed robust global features. PCT reaches the $\text{ER}$ of 13.5\% on transformation corruptions, consistently achieving the best model on all corruption types. In comparison to density and noise corruptions, transformation corruptions are mild and have a minor effect on the local smoothness. Transformer has been demonstrated to have large capacity and a global receptive field, and we believe this design contribute to its resilience to global corruption of point clouds.   

Besides above, we find that SimpleView cannot achieve better robustness under common corruptions than other architectures, despite it high performance on clean data (Table~\ref{tb:total1}), suggesting point cloud-specific designs are indeed desired. Due to its good results on \texttt{Background}, PointNet++ on average performs the best with standard training, achieving an $\text{ER}_\text{cor}$ of 23.6\%, and PCT has a more balanced performance across corruptions, making it the runner-up architecture.

\vspace{-0.2cm}
\subsection{Data Augmentation Strategies}
\label{sec:exp_aug}

As mentioned earlier, we use five additional data augmentation strategies to train the six models. In this section, we examine how these training recipes combined with different models perform on ModelNet40-C, and Table~\ref{tb:aug} presents the overall results. Due to the space limit, we group the evaluation results per corruption patterns and the detailed results are shown in Fig.~\ref{fig:erforda} and ~\ref{fig:erforda2} in Appendix~\ref{ap:experiment}. Several interesting insights can be concluded from our experiments. 

\begin{figure}[t]
\includegraphics[width=0.975\linewidth]{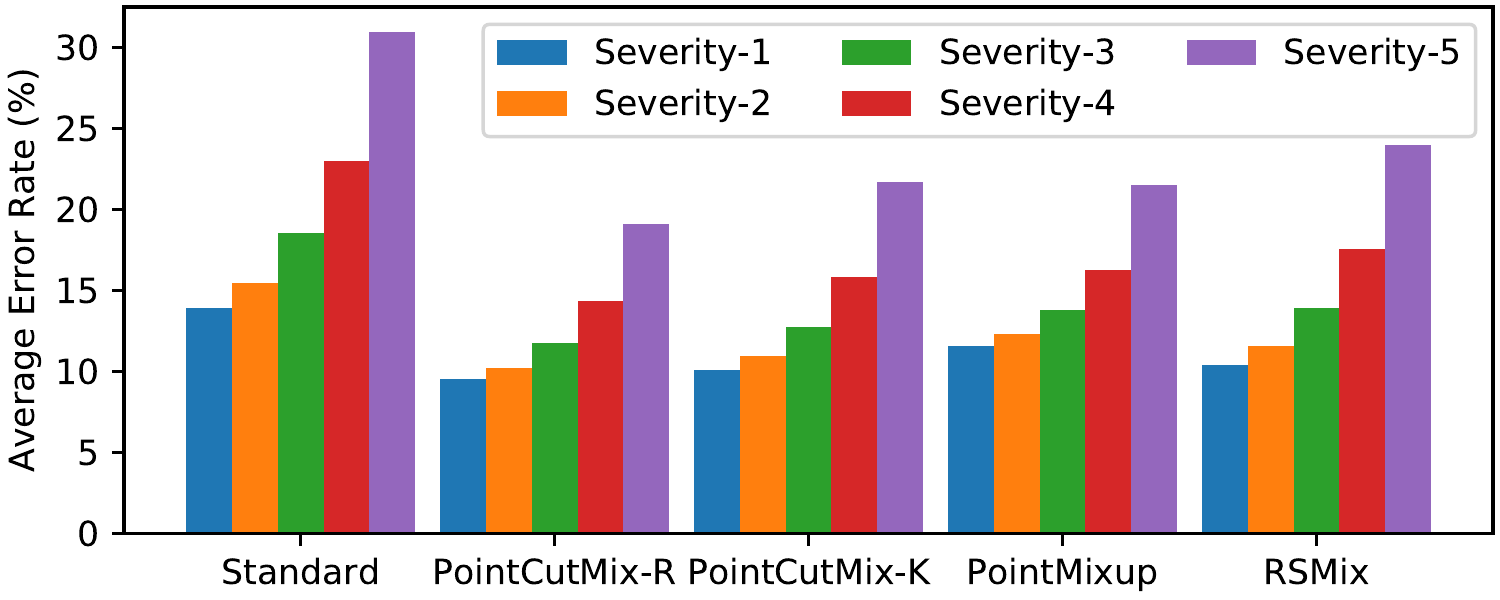}
\vspace{-0.4cm}
\caption{Average Error Rates over Six Models and 13 Corruptions (except for \texttt{Occlusion} and \texttt{LiDAR}). We exclude \texttt{Occlusion} and \texttt{LiDAR} since they do not have different severity levels.}\label{fig:sev}
\vspace{-0.6cm}
\end{figure}

\textbullet~\textit{Insight 3.1} Data augmentation strategies generally improve the corruption robustness of 3D point cloud recognition. 

As Table~\ref{tb:aug} indicates, all the augmentation recipes enhance the overall corruption robustness from 19.6\% to 28.4\%. Data augmentation methods enrich the training set the resulting model more general. Combined with results in Table~\ref{tb:clean}, our paper suggests different conclusions from the original claims: mixing-based augmentations have little gain on clean performance but help generalize to common corruptions.

\textbullet~\textit{Insight 3.2.} No single data augmentation can rule them all. Different augmentation methods have expertise on distinct corruption patterns.

As Table~\ref{tb:aug} presents, PointCutMix-R performs the best on noise corruptions ($\text{ER}=12.7$\%), PointMixup specializes the transformation corruptions ($\text{ER}=14.1$\%), and RSMix is especially robust against density corruptions ($\text{ER}=26.8$\%). Such results also relate to the design of augmentation strategies. In details,  given two point cloud samples $\vx_a$,$\vx_b$ from class $a$ and $b$, PointCutMix-R simply merges ($\oplus$) two randomly selected ($\odot$) subsets together based on hyper-parameter $\lambda$ ($\vx_{aug} = \lambda \odot \vx_a \oplus (1-\lambda)\odot \vx_b$). The two subsets will overlap in the resulting point cloud $\vx_{aug}$. Each point cloud subset can be regarded as a special noise by the other. Thus, it naturally includes noise corruptions with mixing into data augmentations. PointMixup leverages interpolation-based mixing that the transition between two point clouds ($\vx_{aug}= \lambda\vx_a + (1-\lambda)\zeta(\vx_a,\vx_b)$, where $\zeta(\vx_a,\vx_b)$ finds the shortest path for every pair in $\vx_a$ and $\vx_b$). The augmented point cloud is thus locally smooth, which aligns with the transformation corruptions. In contrast, RSMix acts similarly with PointCutMix-K but guarantee a \textit{rigid} mixing of two partial point clouds. There will be no overlaps and each point cloud subset is clustered and isolated in the 3D space. Such patterns correspond to density corruptions in point cloud data.

\textbullet~\textit{Insight 3.3.} Adversarial training does not show superiority on corruption robustness for 3D point cloud recognition.

Adversarial training improves robustness on noise corruptions since we rely on point shifting attacks in the inner maximization stage. \citet{shen2021interpreting} suggest that adversarial rotation training improves the robustness against random rotations. We here motivate future research to present general methods that improve both adversarial and corruption robustness for point cloud learning.

Moreover, with more data augmentation and we find \textbf{PCT} outperforms PointNet++ with augmentation sophisticated training recipes. Such results align with recent studies in corruption robustness of 2D vision tasks as well~\cite{bai2020vitsVScnns}, suggesting the superiority of Transformer-based design in 3D point cloud learning. Surprisingly, the simplest augmentation, PointCutMix-R, achieves the best overall robustness ($\text{ER}_\text{corrup}$=18.7\%). As Fig.~\ref{fig:sev} shows, it is especially helpful on corruptions with high severity levels. We hope our analysis will facilitate future research on designing effective and robust training recipes.

% \begin{figure}[h]
% \includegraphics[height=4cm,width=\linewidth]{figs/placeholder.png}
% \vspace{-0.6cm}
% \caption{Illustration. Samples of different data augmentations}
% \end{figure}

\vspace{-0.25cm}
\subsection{Test-Time Adaptation Methods}
\label{sec:exp_adapt}
\vspace{-0.15cm}

\begin{table}[t]
\scriptsize
\renewcommand\arraystretch{1}
\setlength\tabcolsep{2pt}
\vspace{-0.25cm}
  \caption{\small Error Rates of Different Model Architectures on ModelNet40-C with Different Test-time Adaptation Methods.
%   \AM{Missing number might draw unnecessary criticism. Is there a better way to combine the results?}
  %\BK{Better caption with message/results -- eg see winning hand captions}\AM{Move the detailed result to Appendix and only keep first 5 columns.}
  }
  \label{tb:adapt}
  \centering
\resizebox{1.0\linewidth}{!}{
  \begin{tabular}{l||c|ccc|c|ccc}
    \specialrule{1pt}{1.1pt}{1pt}
    & \multicolumn{4}{c|}{BN} & \multicolumn{4}{c}{TENT} \\
    \cline{2-9}
    %  &CIFAR-10 &\multicolumn{4}{c|}{CIFAR-10-C} &CIFAR-10-$\bar{\text{C}}$ &CIFAR-100 &\multicolumn{4}{c|}{CIFAR-100-C} &CIFAR-100-$\bar{\text{C}}$ &ImageNet & \multicolumn{4}{c|}{ImageNet-C} &ImageNet-$\bar{\text{C}}$ \\
    % \hline
    %Pretext Task     & Parameters  &Clean Accuracy &Adv. Accuracy  &Clean Accuracy &Adv. Accuracy  &Clean Accuracy &Adv. Accuracy\\
    Model (\%) $\downarrow$ & $\text{ER}_\text{cor}$ & Density & Noise & Trans. & $\text{ER}_\text{cor}$ & Density & Noise & Trans. \\
\noalign{\global\arrayrulewidth1pt}\hline\noalign{\global\arrayrulewidth0.4pt}     
PointNet   & 25.9 & 26.8 & 29.6 & 21.2 & 27.3 & 27.8 & 31.0 & 23.0 \\
PointNet++ & \textbf{16.9} & \textbf{24.2} & \textbf{12.7} & 13.8 & \textbf{16.7} & 24.1 & \underline{12.3} & 13.6 \\
DGCNN      & 21.1 & 31.4 & 19.5 & \underline{12.5} & 20.9 & 30.7 & 19.4 & \underline{12.5} \\
RSCNN      & 20.0 & \underline{26.4} & 16.7 & 17.0 & 19.4 & \textbf{26.0} & 15.9 & 16.4 \\
PCT        & \underline{19.5} & 27.9 & 18.2 & \textbf{12.4} & 18.0 & \underline{26.9} & 15.4 & \textbf{11.7} \\
SimpleView & 19.8 & 29.8 & \underline{13.8} & 15.8 & \underline{16.9} & 28.1 &  \textbf{9.9} & 12.8 \\
\hline
Average    & 20.5 & 27.7 & 18.4 & 15.4 & 19.9 & 27.3 & 17.3 & 15.0 \\
\noalign{\global\arrayrulewidth1pt}\hline\noalign{\global\arrayrulewidth0.4pt}
  \end{tabular}
    }
  \vspace{-0.7cm}
\end{table}

Besides the introduced training-time strategies, we evaluate test-time adaptation methods on our ModelNet40-C. Specifically, we for the first time adapt the BN~\cite{schneider2020improving} and TENT~\cite{wang2020tent} to point cloud recognition. BN updates the statistics of BatchNorm~\cite{ioffe2015batch} layers (\textit{i.e.}, $\mu$ and $\sigma$.) based on the incoming batch of testing point clouds. TENT updates both the statistics and weight parameters (\textit{i.e.}, $\gamma$ and $\beta$.) of BatchNorm layers to minimize the cross-entropy of the output layer. Table~\ref{tb:adapt} presents the evaluation results. 

\textbullet~\textit{Insight 4.1.} Test-time adaptations overall perform worse than data augmentation strategies in improving robustness.

BN and TENT consistently help enhance the corruption robustness for point cloud recognition. However, we find they overall are not as effective as data augmentation strategies. Such observations are different from 2D vision tasks, and we attribute the reason to the nature of corruptions in 3D space. Corruption robustness benchmarks for 2D images are created by changing the RGB values. Corruptions in the 3D space directly modify both the numbers and positions of points. The distributional shift is thus large between corrupted and clean point clouds.

% \begin{figure}[h]
% \includegraphics[height=4cm,width=\linewidth]{figs/placeholder.png}
% \vspace{-0.6cm}
% \caption{show how detailed tent and bn helps on three corruptions}
% \end{figure}

\textbullet~\textit{Insight 4.2.} Test-time adaptation is surprisingly useful on tough corruptions.

We find that TENT on average helps achieve the strongest robustness on \texttt{Occlusion} (ER=47.6\%), \texttt{LiDAR} (ER=54.1\%), and \texttt{Rotation} (ER=19.8\%) corruptions, outperforming the best augmentation method by 6.7\%, 1.9\%, and 7.9\% respectively. Especially, we find test-time adaptation methods achieve the best $\text{ER}_{\text{rotation},5} =35.6\%$, which is a 27.1 \% improvement over the best augmentation strategy. Augmentation strategies cannot handle these difficult corruptions, but test-time adaptation methods deliver a more consistent improvement.

\begin{table}[h]
\scriptsize
\renewcommand\arraystretch{1.1}
\setlength\tabcolsep{2pt}
\vspace{-0.5cm}
  \caption{\small Error Rates of Different Model Architectures on ModelNet40-C with PointCutMix-R and TENT.
%   \AM{Missing number might draw unnecessary criticism. Is there a better way to combine the results?}
  %\BK{Better caption with message/results -- eg see winning hand captions}\AM{Move the detailed result to Appendix and only keep first 5 columns.}
  }
  \label{tb:mega}
  \centering
\resizebox{1.0\linewidth}{!}{
  \begin{tabular}{l||cccccc}
    \specialrule{1pt}{1.1pt}{1pt}
    %  &CIFAR-10 &\multicolumn{4}{c|}{CIFAR-10-C} &CIFAR-10-$\bar{\text{C}}$ &CIFAR-100 &\multicolumn{4}{c|}{CIFAR-100-C} &CIFAR-100-$\bar{\text{C}}$ &ImageNet & \multicolumn{4}{c|}{ImageNet-C} &ImageNet-$\bar{\text{C}}$ \\
    % \hline
    %Pretext Task     & Parameters  &Clean Accuracy &Adv. Accuracy  &Clean Accuracy &Adv. Accuracy  &Clean Accuracy &Adv. Accuracy\\
    Corruption (\%) $\downarrow$ & PointNet & PointNet++ & DGCNN & RSCNN & PCT & SimpleView \\
\noalign{\global\arrayrulewidth1pt}\hline\noalign{\global\arrayrulewidth0.4pt}     
Density & 24.8 & \textbf{21.8} & 24.7 & \underline{22.7} & 22.8 & 23.5 \\
Noise & 15.7 & 9.5 & \underline{9.2} & 10.7 & \textbf{9.0} & 9.7 \\
Trans. & 15.2 & 12.1 & \underline{10.4} & 12.9 & \textbf{10.1} & 13.3 \\
\hline
$\text{ER}_\text{cor}$ & 18.5 & \underline{14.5} & 14.8 & 15.4 & \textbf{13.9} & 15.5 \\
\noalign{\global\arrayrulewidth1pt}\hline\noalign{\global\arrayrulewidth0.4pt}
  \end{tabular}
    }
  \vspace{-0.2cm}
\end{table}

We have so far demonstrated that PointCutMix-R and TENT obtain the best among the training- and test-time methods, in terms of the overall $\text{ER}_\text{cor}$. We here evaluate the performance of the combination of PointCutMix-R and TENT as they do not conflict with each other. As presented in Table~\ref{tb:mega}, we find that the combined solution further improves the corruption robustness by 14.7\%. To our best knowledge, there is no test-time adaptation designs specific for point cloud learning, and we hope our study will shed light on future research on corruption robustness in this area.

\vspace{-0.25cm}
\section{Discussion and Conclusion}
\vspace{-0.1cm}

Through our systematic benchmarking and analysis, we found that the performance discrepancies of different point cloud recognition models across different corruptions are much larger than 2D architectures. This suggests future studies on a universal architecture design for 3D point cloud to be a worthwhile direction. In the future, we plan to extend our current benchmark to complex tasks like point cloud segmentation and object detection to facilitate future research on robustness in the 3D domain.

To conclude, we have presented ModelNet40-C, the first comprehensive benchmark for corruption robustness of point cloud recognition models. We have unveiled the massive performance degradation on our ModelNet40-C for six representative models. We also provided critical insights on how different architecture and data augmentation designs affect model robustness on different corruptions. For example, Transformer appears to be a promising architecture in improving robustness in 3D vision tasks. Our study on test-time adaptation in point cloud recognition shows its potential as a robustness strategy. We hope that our ModelNet40-C benchmark will benefit future research in developing robust 3D point cloud models.
% \input{sections/conclusion}

\clearpage
\clearpage

{\small
\bibliography{example_paper}
\bibliographystyle{icml2022}
}

%%%%%%%%%%%%%%%%%%%%%%%%%%%%%%%%%%%%%%%%%%%%%%%%%%%%%%%%%%%%%%%%%%%%%%%%%%%%%%%
%%%%%%%%%%%%%%%%%%%%%%%%%%%%%%%%%%%%%%%%%%%%%%%%%%%%%%%%%%%%%%%%%%%%%%%%%%%%%%%
% APPENDIX
%%%%%%%%%%%%%%%%%%%%%%%%%%%%%%%%%%%%%%%%%%%%%%%%%%%%%%%%%%%%%%%%%%%%%%%%%%%%%%%
%%%%%%%%%%%%%%%%%%%%%%%%%%%%%%%%%%%%%%%%%%%%%%%%%%%%%%%%%%%%%%%%%%%%%%%%%%%%%%%
\newpage
\appendix
\onecolumn
\section{ModelNet40-C}
\label{ap:modelnet}
We elaborate the creation of ModelNet40-C in this section. The detailed implementation can be found in our codebase, which is included in the supplementary materials.

\begin{figure}[h]
\centering
\includegraphics[width=0.75\linewidth]{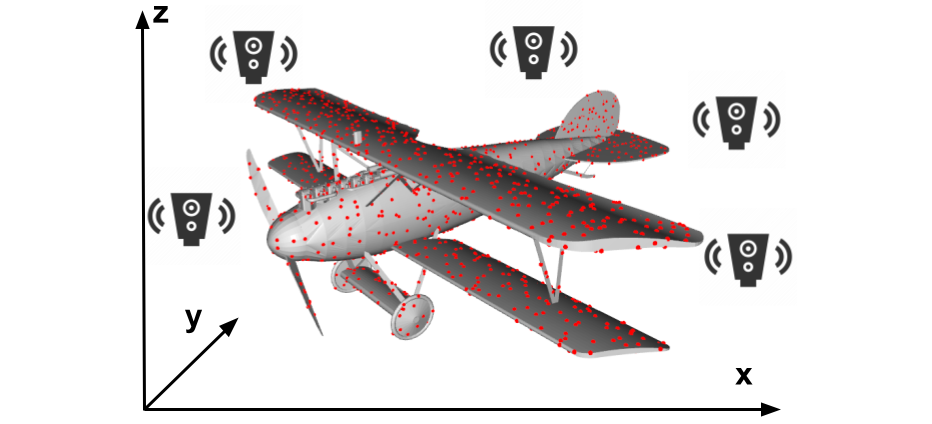}\label{fig:ap_demo}
\vspace{-0.2cm}
\caption{Illustration of \texttt{Occlusion} and \texttt{LiDAR} Corruption Generation.}
\vspace{-0.4cm}
\end{figure}

\texttt{Occlusion} and \texttt{LiDAR} share similar general corruption features. We leverage five viewing angles to construct these two corruptions on ModelNet40, as shown in Fig.~\ref{fig:ap_demo}. Specifically, we utilize ray tracing algorithms on the original meshed from ModelNet40 to generate the point cloud. Let the facing direction of the object as $0^\circ$ pivoting the $z$ axis, we use $0^\circ$, $72^\circ$, $144^\circ$, $216^\circ$, and $288^\circ$ as our viewing angles, the viewing angles between the $xy$ plane are randomly sampled from is $30^\circ-60^\circ$. For \texttt{LiDAR}, we additionally render the generated point cloud into the vertically multi-line style to simulate the pattern of the LiDAR sensor.

For \texttt{Local\_Density\_Inc} and \texttt{Local\_Density\_Dec}, we first sample a number of anchor points based the severity level. We further find the $k$NN of the anchor points and up-sample or down-sample them to increase and decrease their local density, respectively.
Similarly, \texttt{Cutout} discards the full $k$NN ($k=50$) subsets of the anchor points to simulate the sensor limitations of LiDAR and other scanning devices.

\texttt{Gaussian} and \texttt{Uniform} noises are sampled from Gaussian and uniform distributions with different $\sigma$ and $\epsilon$ based on the severity level. For the \texttt{Background} noise, we randomly sample different numbers of points in the edge-length-2 cube that bounds the point cloud based on the severity level. For \texttt{Impulse} noise, we first sample different numbers of points based on the severity level and assign the maximum magnitude of perturbation $\ell_\infty = 0.05$ to them. For the \texttt{Upsampling} noise, we first choose different numbers of points based on the severity level and generate new points around the selected anchors, bounded by $\ell_\infty = 0.05$.

For \texttt{Rotation} and \texttt{Shear}, we have introduced their construction in \S~\ref{sec:cor}. As mentioned, we allow relatively small transformations since we find larger ones will affect the human perception of the object class as well.

For deformation-based corruptions \texttt{FFD}, \texttt{RBF}, and \texttt{Inv\_RBF}, we assign 5 control points along each $xyz$ axis, resulting in 125 control points in total. We choose the deformation distance based on the severity level and randomly assign their directions in the 3D space. The deformations then are formulated based on the interpolation functions that we choose in~\S~\ref{sec:cor}.

\begin{figure}[h]
\centering
\includegraphics[width=0.95\linewidth]{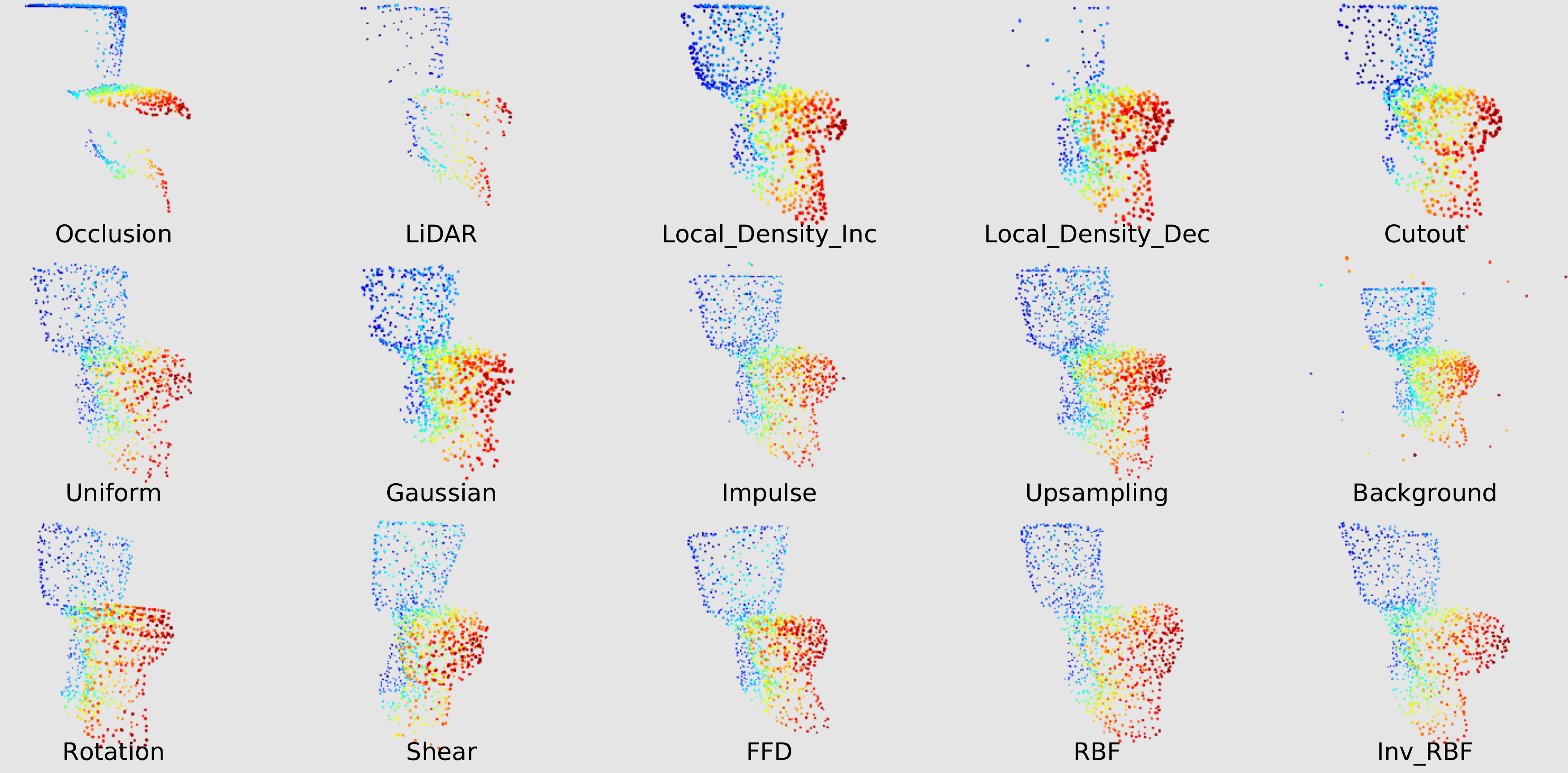}
\vspace{-0.2cm}
\caption{Visualization of Samples from ModelNet40-C - ``Toliet'' Class.}\label{fig:ap_vis1}
\vspace{-0.4cm}
\end{figure}

\begin{figure}[h]
\centering
\includegraphics[width=0.95\linewidth]{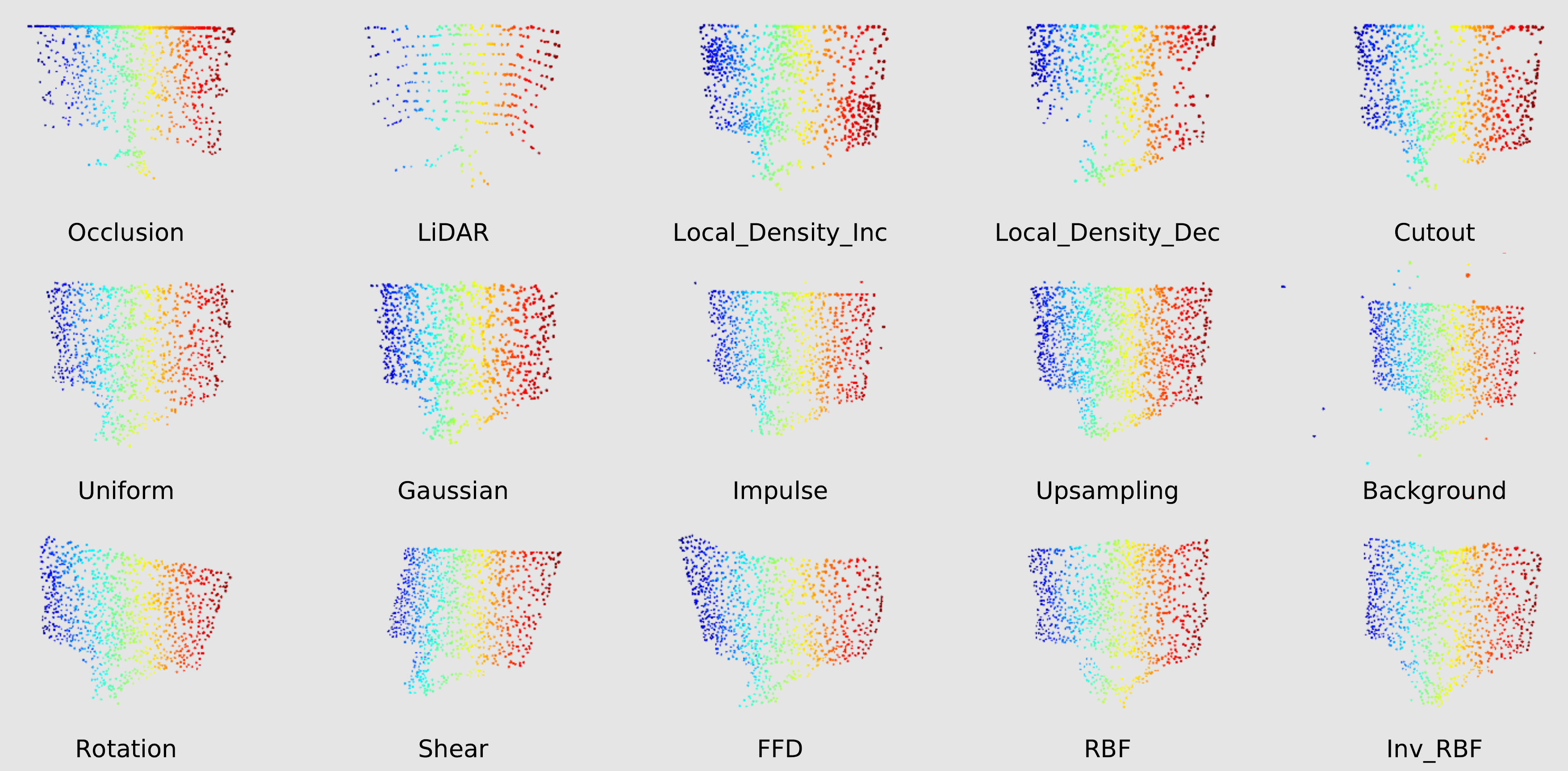}
\vspace{-0.2cm}
\caption{Visualization of Samples from ModelNet40-C - ``Monitor'' Class.}\label{fig:ap_vis2}
\vspace{-0.4cm}
\end{figure}

We visualize two additional groups of sample point clouds from ModelNet40-C in Fig.~\ref{fig:ap_vis1}. and~\ref{fig:ap_vis2}.

\section{Experiments}
\label{ap:experiment}

\subsection{Data Augmentation Setups}

We introduce the detailed setting of our experiments and analysis in this section. For all mixing-based data augmentation strategies, we have a hyper-parameter $\lambda$ to determine the weight of two samples to mix, as well as the weight of the virtual label vector:
\begin{equation}
   y_{aug} = \lambda \cdot y_a + (1-\lambda) \cdot y_b
\end{equation}
where $y_a$ and $y_b$ are the label vectors for point cloud $\vx_a$ and $\vx_b$. We set $\lambda = 0.5$, which has shown to achieve good results reported in~\cite{zhang2021pointcutmix,chen2020pointmixup,lee2021regularization}. For adversarial training, we use the point shifting attack in the adversarial inner maximization:
\begin{equation}
    \vx^{s+1} = \Pi_{\vx + \mathcal{S}}(\vx^{s} + \alpha \cdot \mathrm{sign}(\nabla_{\vx^{s}} \mathcal{L}(\vx^s,y; f)));\quad \vx^0 = \vx + \mathrm{U}(-\epsilon,\epsilon)
    \label{eq:pgd}
\end{equation}
where $\epsilon = 0.05$, $\alpha = 0.01$ and we use seven steps PGD, as suggested by~\citet{sun2021adversarially}.

\subsection{Evaluation Results}

\begin{figure*}[h]
\centering
\includegraphics[width=0.975\linewidth]{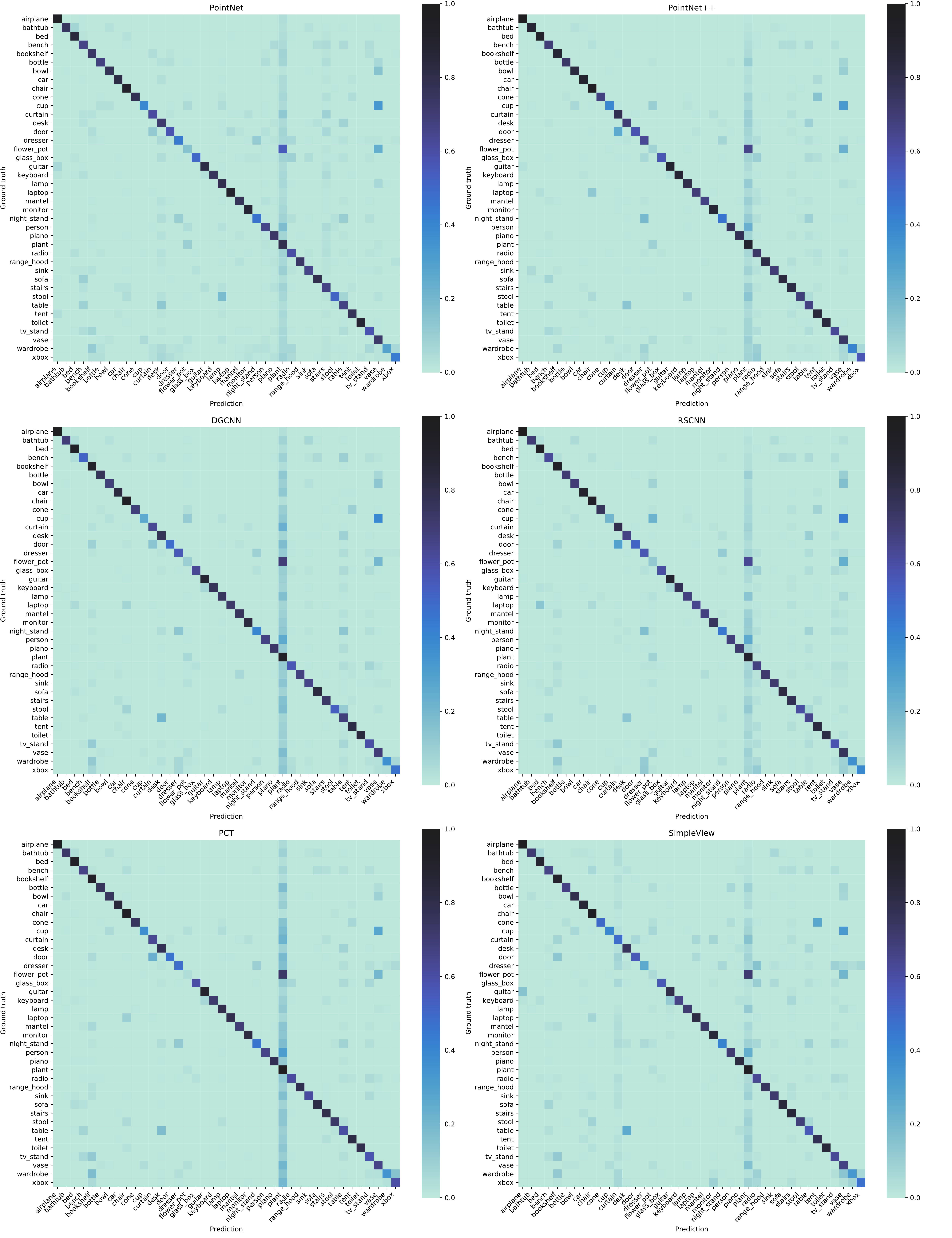}
\caption{Confusion Matrices for Each Model on ModelNet40-C with Standard Training.}
\label{fig:apconfusion}
\end{figure*}

\begin{figure*}[t]
\centering
\includegraphics[width=\linewidth]{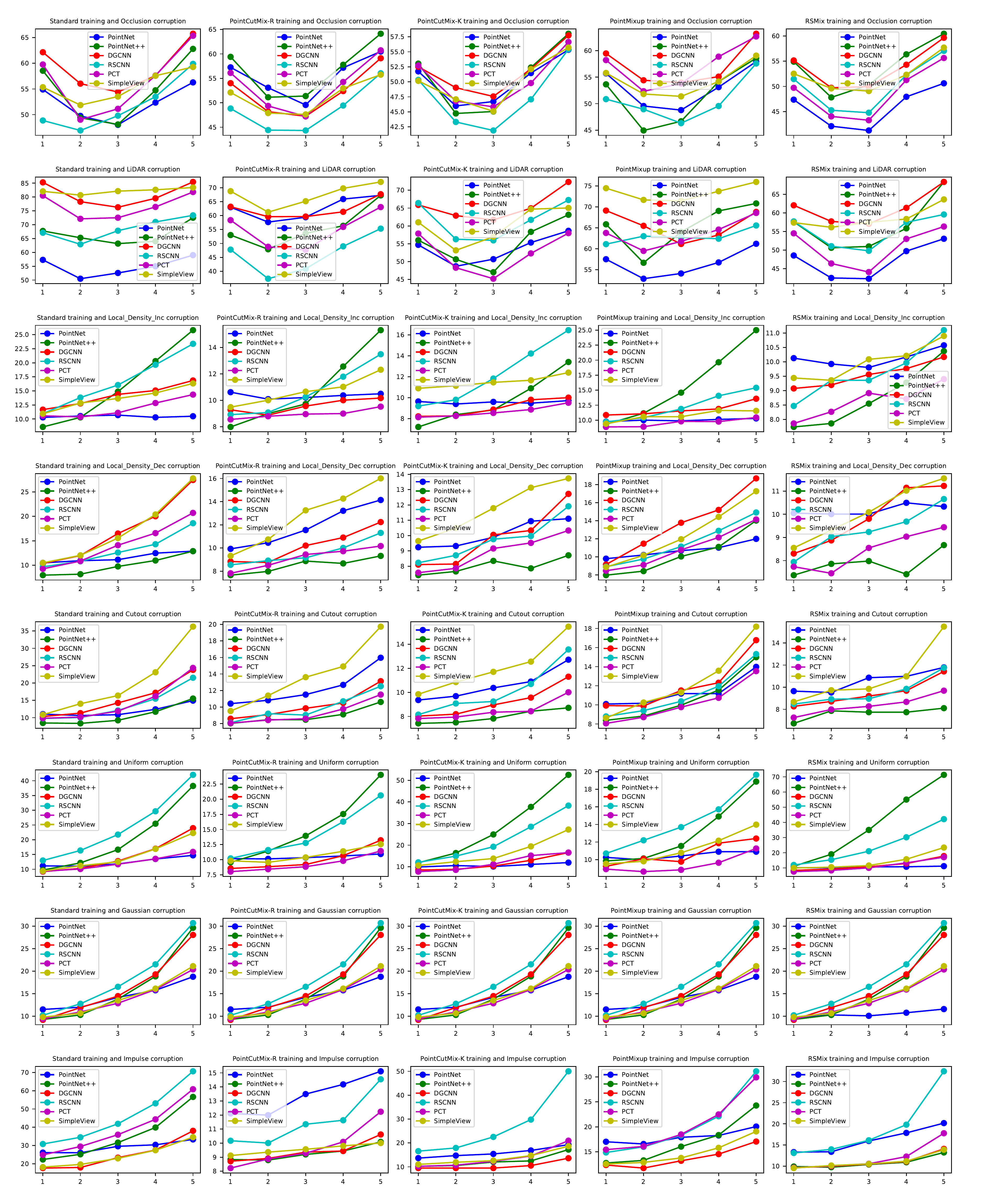}
\vspace{-0.5cm}
\caption{Error Rates of Different Models with Different Data Augmentation Strategies on ModelNet40-C.}
\label{fig:erforda}
\vspace{-0.4cm}
\end{figure*}

\begin{figure*}[t]
\centering
\includegraphics[width=\linewidth]{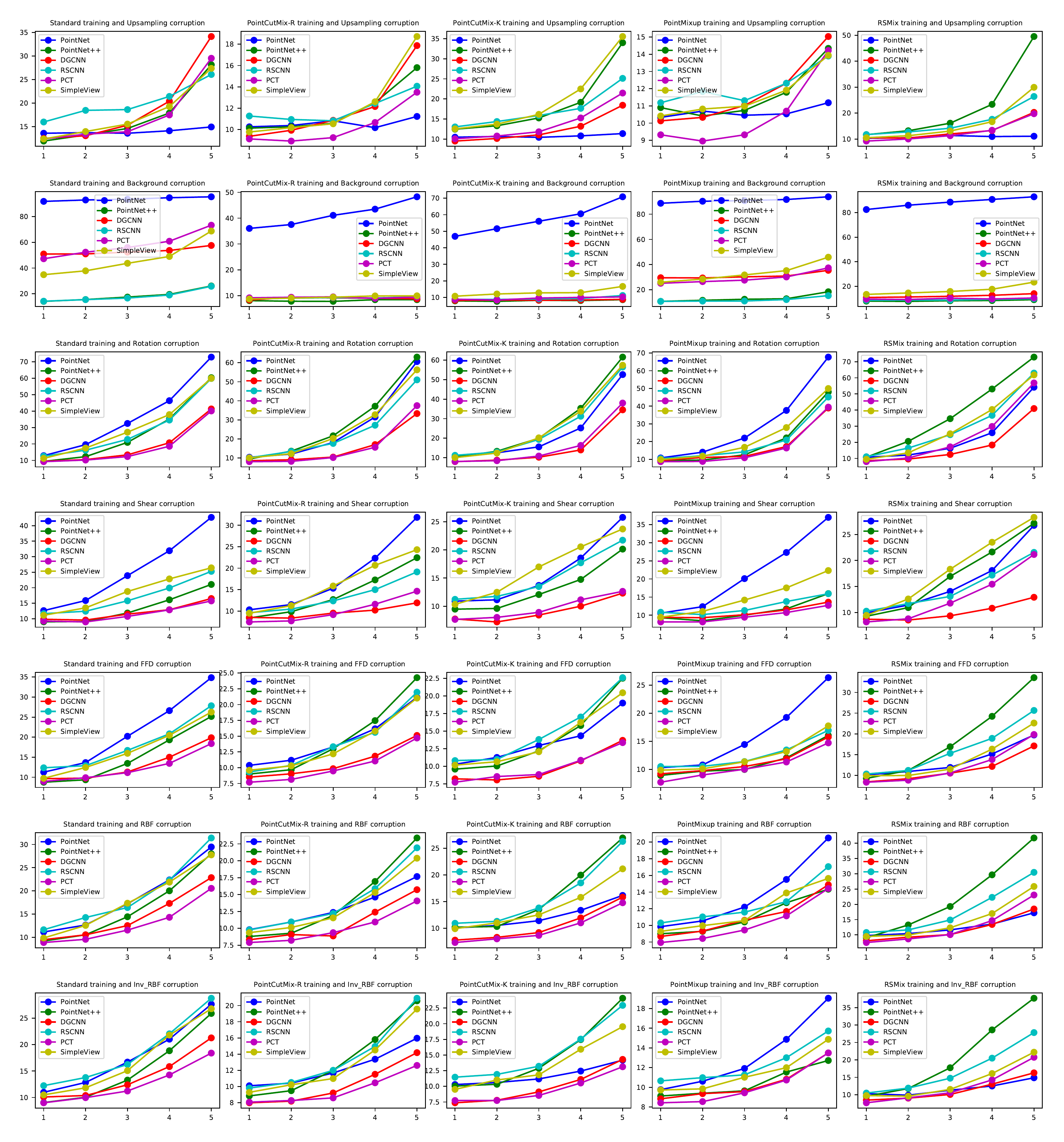}
\vspace{-0.5cm}
\caption{Error Rates of Different Models with Different Data Augmentation Strategies on ModelNet40-C (Cont'd).}
\label{fig:erforda2}
\vspace{-0.4cm}
\end{figure*}

\begin{figure*}[t]
\centering
\includegraphics[width=\linewidth]{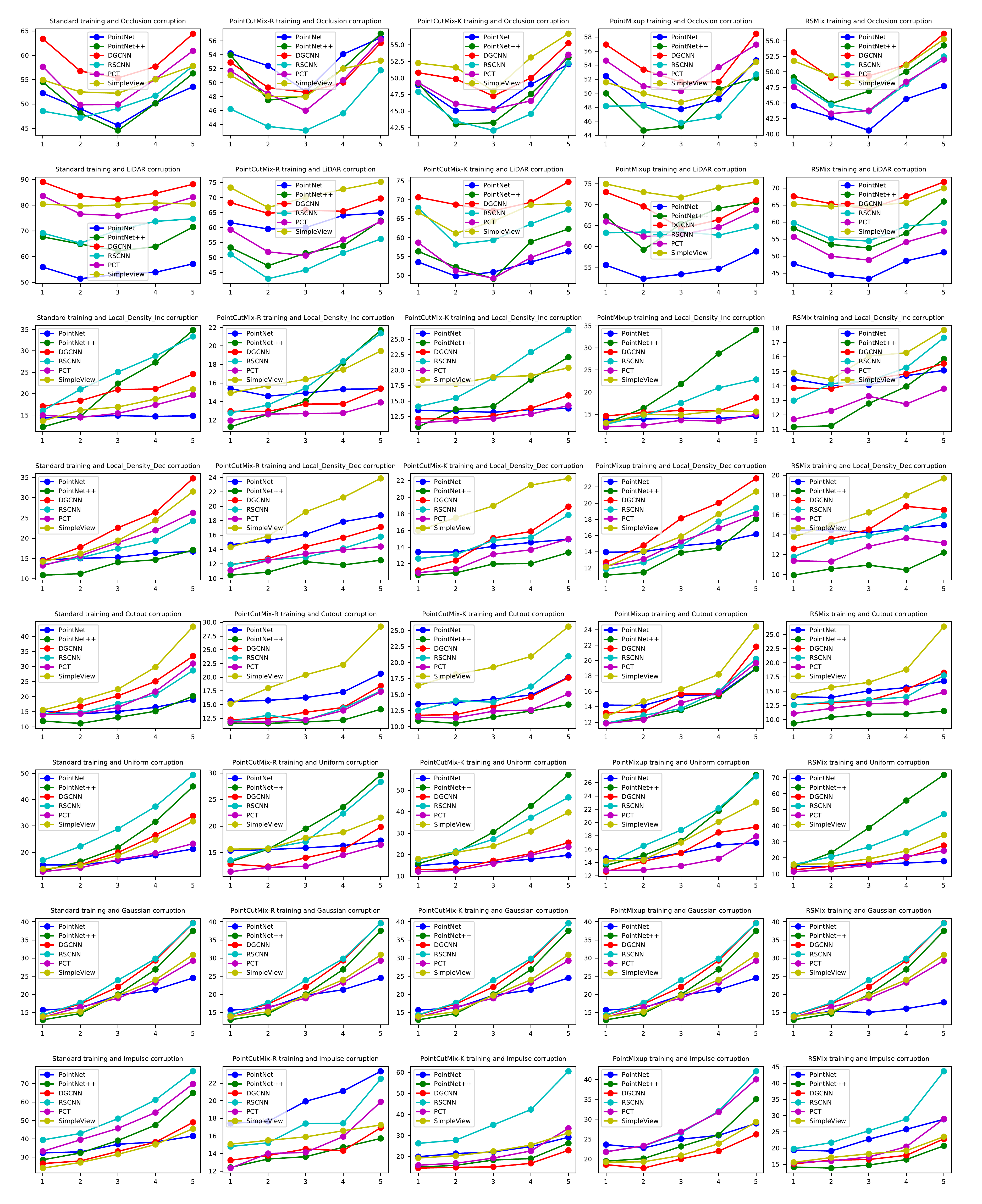}
\vspace{-0.5cm}
\caption{Class-wise Mean Error Rates of Different Models with Different Data Augmentation Strategies on ModelNet40-C.}
\label{fig:merforda}
\vspace{-0.4cm}
\end{figure*}

\begin{figure*}[t]
\centering
\includegraphics[width=\linewidth]{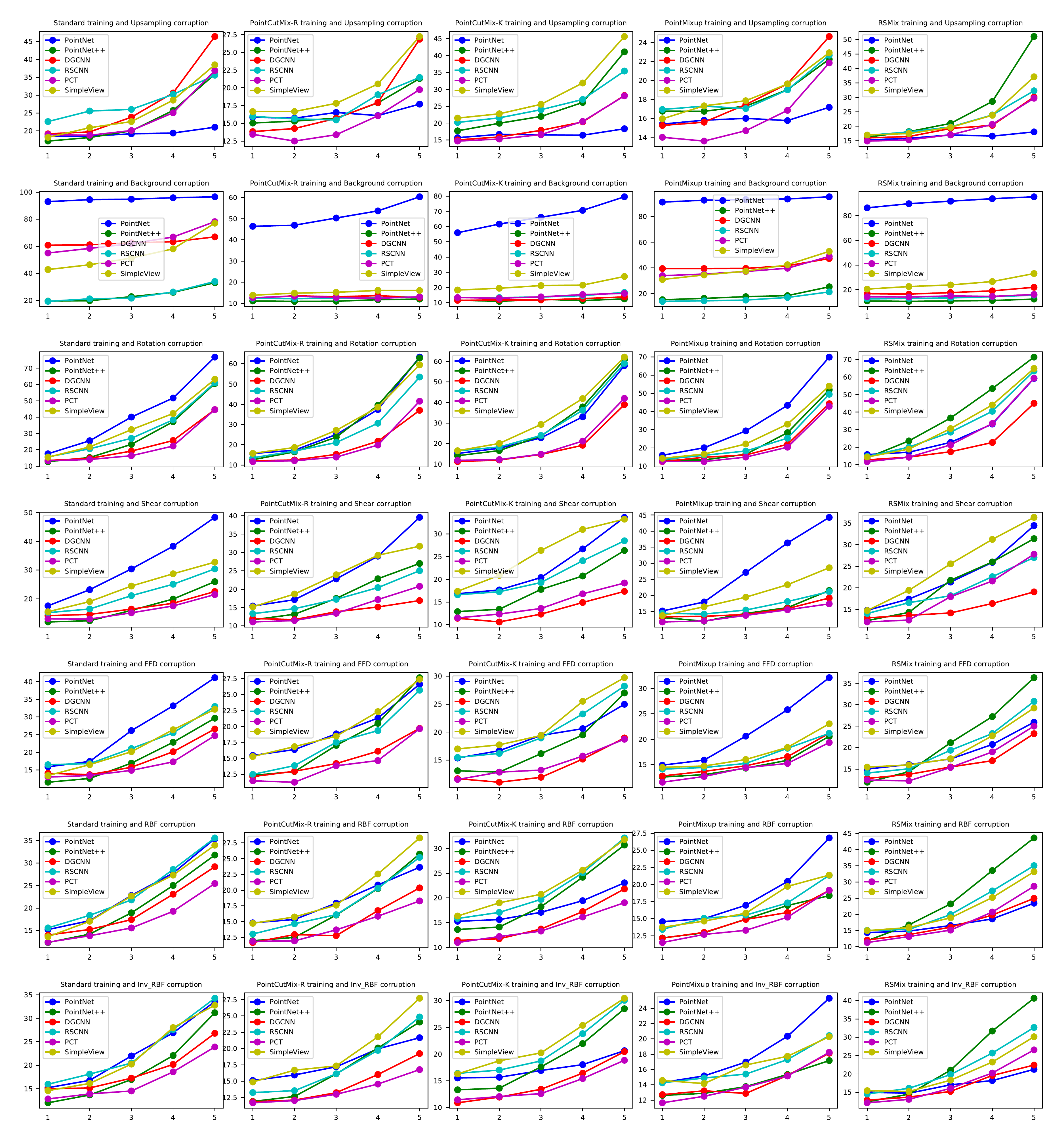}
\vspace{-0.5cm}
\caption{Class-wise Mean Error Rates of Different Models with Different Data Augmentation Strategies on ModelNet40-C (Cont'd).}
\label{fig:merforda2}
\vspace{-0.4cm}
\end{figure*}

\begin{figure*}[t]
\centering
\includegraphics[width=0.6\linewidth]{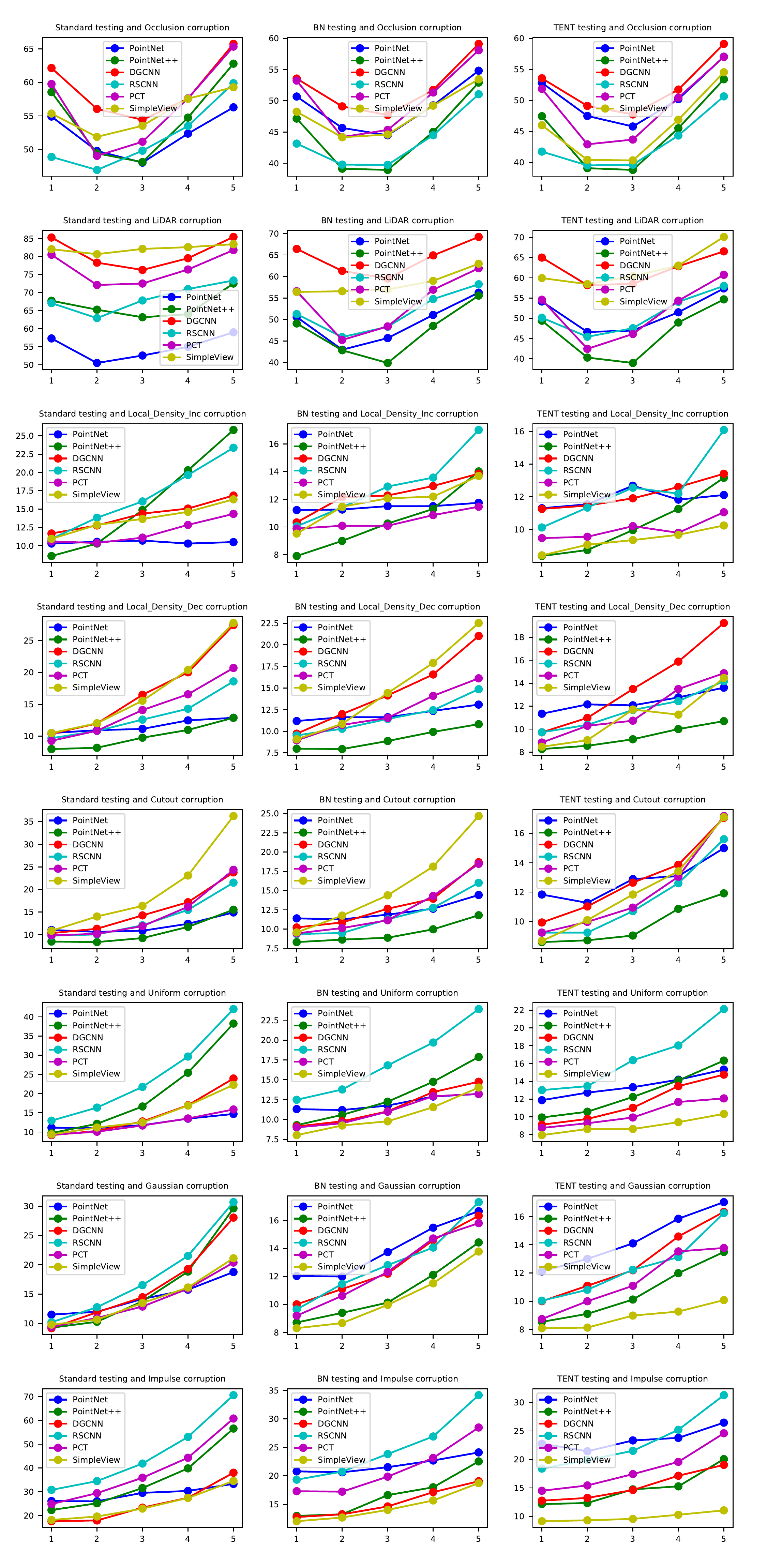}
\vspace{-0.5cm}
\caption{Error Rates of Different Models with Different Test-time Adaptation Methods on ModelNet40-C.}
\label{fig:erforadapt}
\vspace{-0.4cm}
\end{figure*}

\begin{figure*}[t]
\centering
\includegraphics[width=0.6\linewidth]{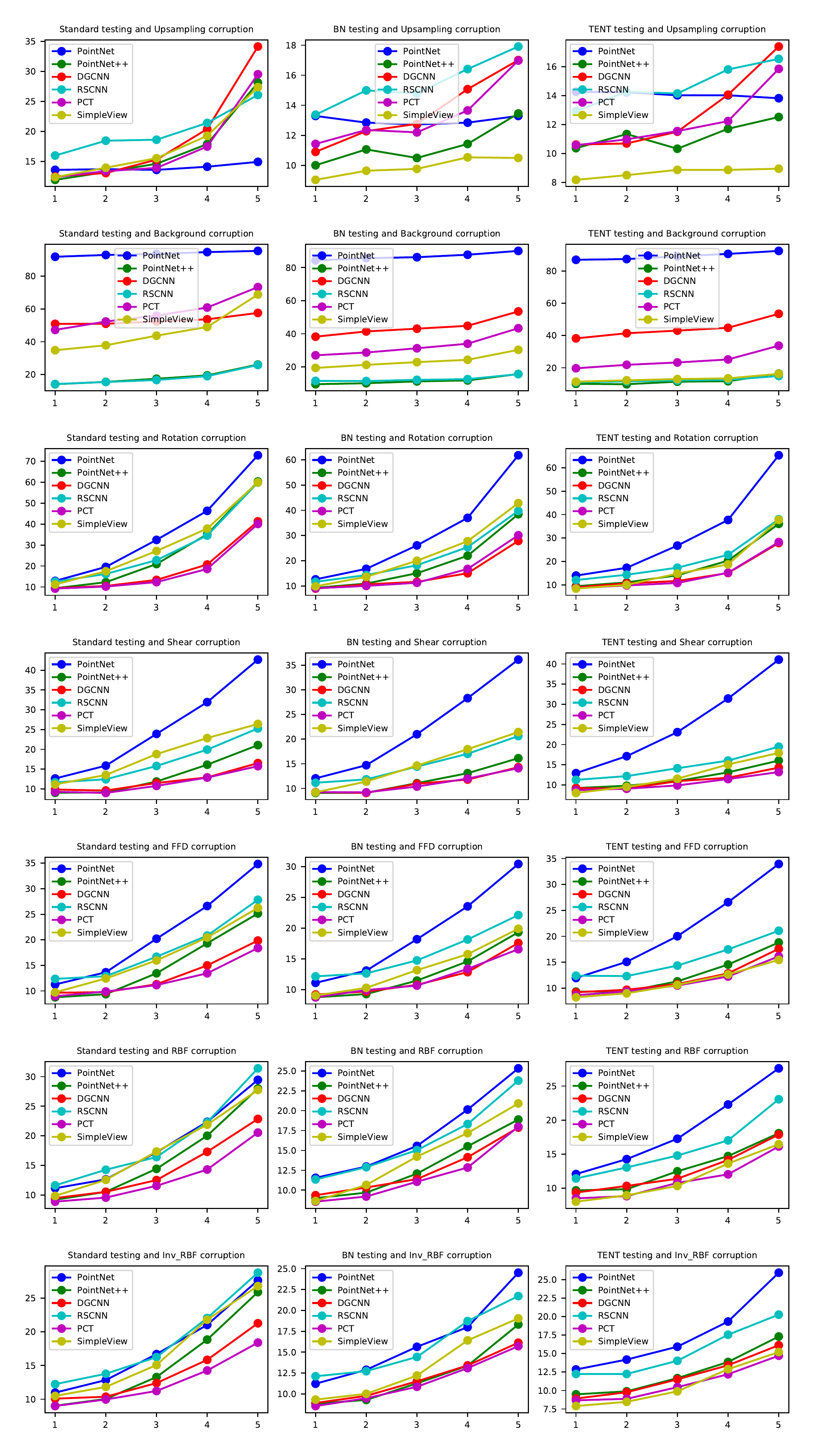}
\vspace{-0.5cm}
\caption{Error Rates of Different Models with Different Test-time Adaptation Methods on ModelNet40-C (Cont'd).}
\label{fig:erforadapt2}
\vspace{-0.4cm}
\end{figure*}

\begin{figure*}[t]
\centering
\includegraphics[width=0.6\linewidth]{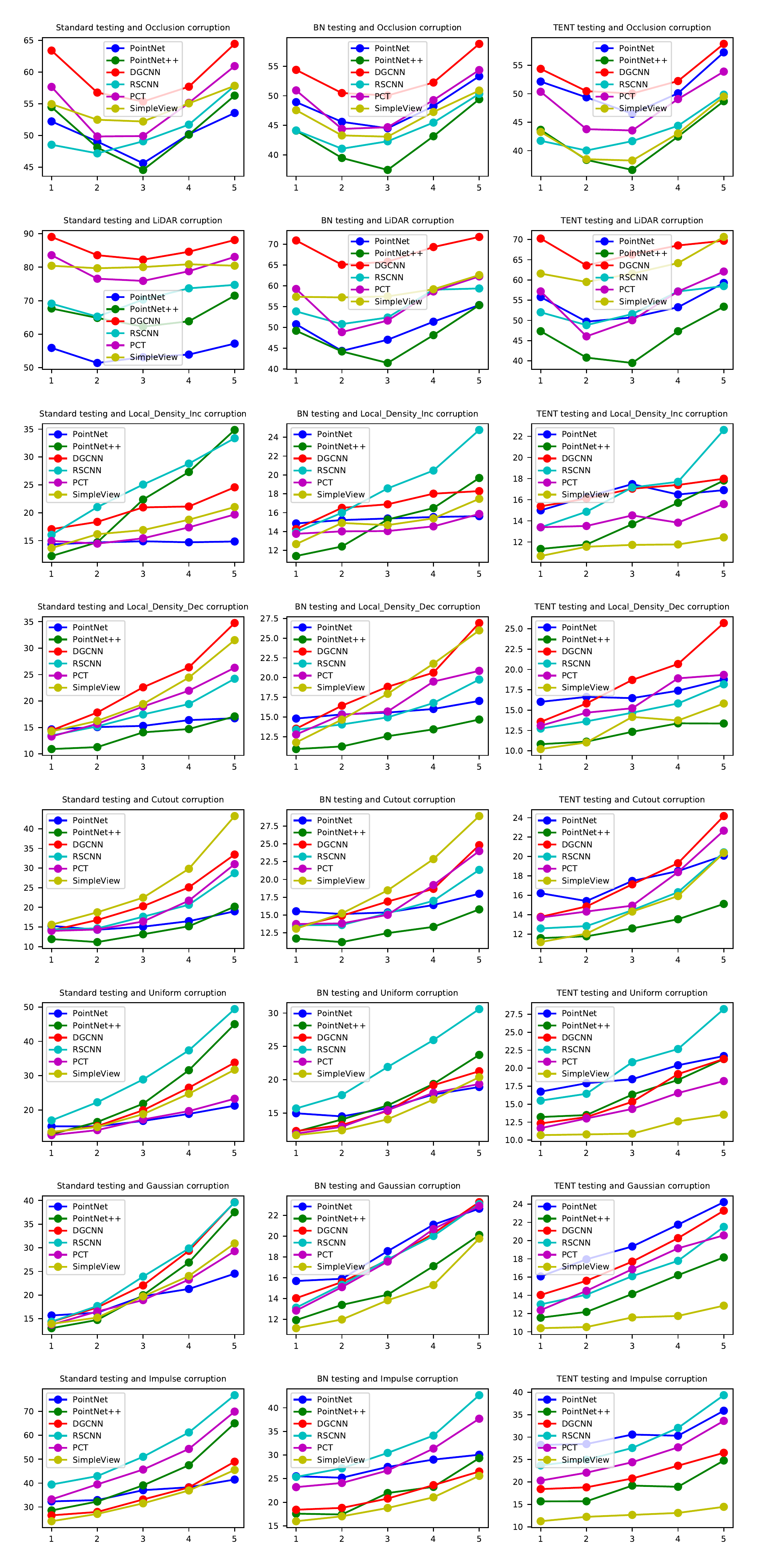}
\vspace{-0.5cm}
\caption{Class-wise Mean Error Rates of Different Models with Different Test-time Adaptation Methods on ModelNet40-C.}
\label{fig:merforadapt}
\vspace{-0.4cm}
\end{figure*}

\begin{figure*}[t]
\centering
\includegraphics[width=0.6\linewidth]{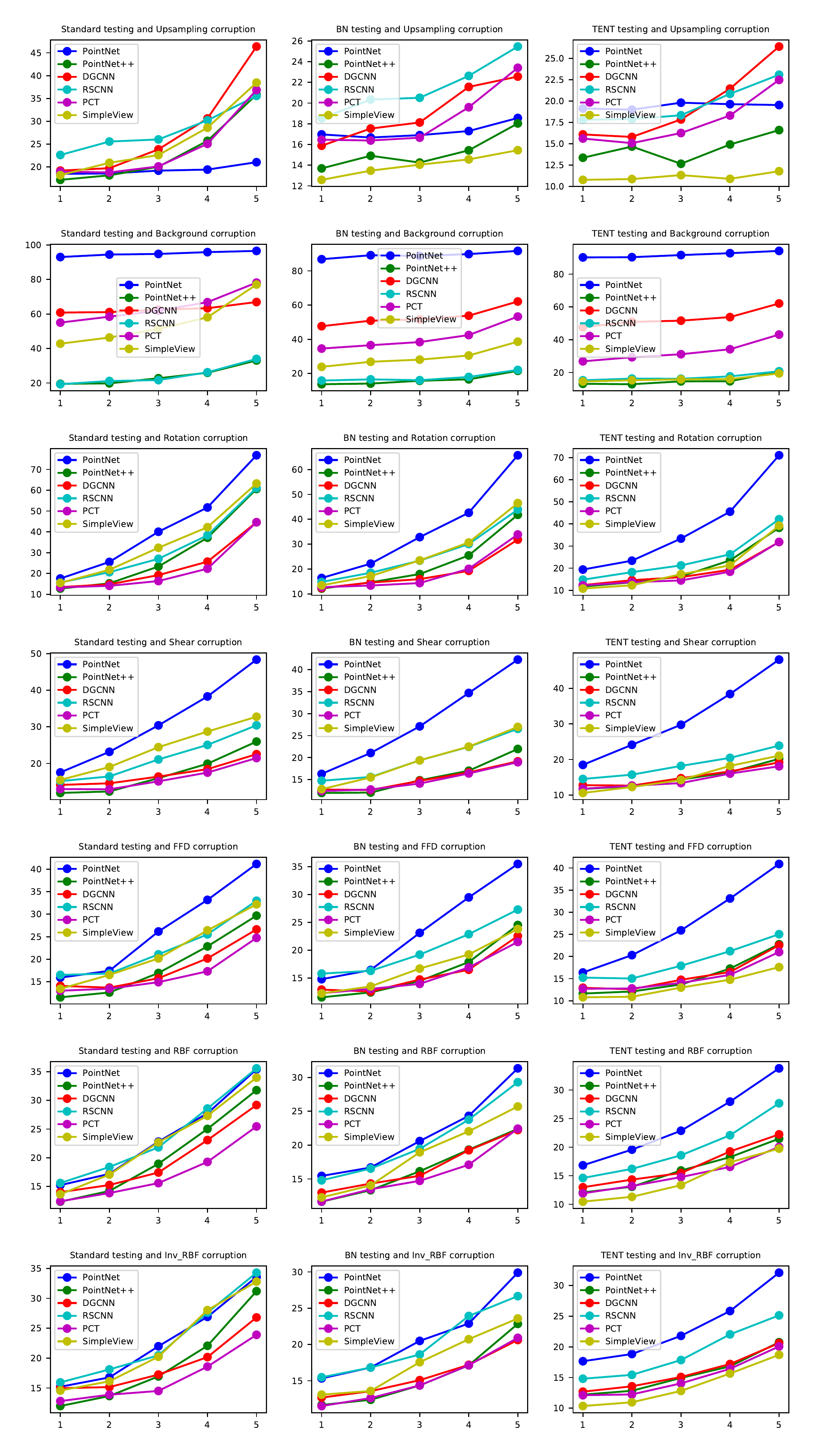}
\vspace{-0.5cm}
\caption{Class-wise Mean Error Rates of Different Models with Different Test-time Adaptation Methods on ModelNet40-C (Cont'd).}
\label{fig:merforadapt2}
\vspace{-0.4cm}
\end{figure*}

We illustrate the confusion matrices of all six models with standard training in Fig.~\ref{fig:apconfusion} to show the validity of ModelNet40-C. We present our detailed evaluation results containing 3,180 data points. Fig.~\ref{fig:erforda} and~\ref{fig:erforda2} shows the model comparison on all data augmentation strategies. The class-wise mean error rates (mER) are shown in Figure~\ref{fig:merforda} and~\ref{fig:merforda2}. The detailed results of test-time adaptation methods (ER and mER) are shown in Fig.~\ref{fig:erforadapt},~\ref{fig:erforadapt2},~\ref{fig:merforadapt}, and~\ref{fig:merforadapt2}.

%%%%%%%%%%%%%%%%%%%%%%%%%%%%%%%%%%%%%%%%%%%%%%%%%%%%%%%%%%%%%%%%%%%%%%%%%%%%%%%
%%%%%%%%%%%%%%%%%%%%%%%%%%%%%%%%%%%%%%%%%%%%%%%%%%%%%%%%%%%%%%%%%%%%%%%%%%%%%%%

\end{document}